\documentclass[acmtog, review=false]{acmart}
\AtBeginDocument{%
  \providecommand\BibTeX{{%
    \normalfont B\kern-0.5em{\scshape i\kern-0.25em b}\kern-0.8em\TeX}}}

\setcopyright{acmcopyright}
\copyrightyear{2018}
\acmYear{2018}
\acmDOI{XXXXXXX.XXXXXXX}

\acmConference[Conference acronym 'XX]{Make sure to enter the correct
  conference title from your rights confirmation emai}{June 03--05,
  2018}{Woodstock, NY}
\acmPrice{15.00}
\acmISBN{978-1-4503-XXXX-X/18/06}

\usepackage{color}
\usepackage{graphicx}
\usepackage{subfigure}
\usepackage{float}
\usepackage{xcolor}
\usepackage{multirow}
\usepackage{bm}

\usepackage{amsfonts,amssymb,amsmath,amsthm}
\usepackage{bbding}
\usepackage{pifont}
\usepackage{wasysym}

\begin{document}

\title{StyleHEAT: One-Shot High-Resolution Editable Talking Face Generation via Pre-trained StyleGAN}

\author{Fei Yin}
\affiliation{%
  \institution{Tsinghua Shenzhen International Graduate School, Tsinghua University, China}
}

\author{Yong Zhang}
\affiliation{
  \institution{Tencent AI Lab, China}
}

\author{Xiaodong Cun}
\affiliation{
  \institution{Tencent AI Lab, China}
}

\author{Mingdeng Cao}
\affiliation{
  \institution{Tsinghua Shenzhen International Graduate School, Tsinghua University, China}
}

\author{Yanbo Fan}
\affiliation{
  \institution{Tencent AI Lab, China}
}

\author{Xuan Wang}
\affiliation{
  \institution{Tencent AI Lab, China}
}

\author{Qingyan Bai}
\affiliation{
  \institution{Tsinghua Shenzhen International Graduate School, Tsinghua University, China}
}

\author{Baoyuan Wu}
\affiliation{
  \institution{The Chinese University of Hong Kong, Shenzhen, China}
}

\author{Jue Wang}
\affiliation{
  \institution{Tencent AI Lab, China}
}

\author{Yujiu Yang}
\affiliation{
  \institution{Tsinghua Shenzhen International Graduate School, Tsinghua University, China}
}
\renewcommand{\shortauthors}{Trovato and Tobin, et al.}


\begin{abstract}

%

One-shot talking face generation aims at synthesizing a high-quality talking face video from an arbitrary portrait image, driven by a video or an audio segment. 
One challenging quality factor is the resolution of the output video: higher resolution conveys more details. 
In this work, we investigate the latent feature space of a pre-trained StyleGAN and discover some excellent spatial transformation properties. 
Upon the observation, we explore the possibility of using a pre-trained StyleGAN to break through the resolution limit of training datasets. 
We propose a novel unified framework based on a pre-trained StyleGAN that enables a set of powerful functionalities, \textit{i.e.,} \textit{high-resolution video generation, disentangled control by driving video or audio, and flexible face editing}. 
Our framework elevates the resolution of the synthesized talking face to 1024$\times$1024 for the first time, even though the training dataset has a lower resolution. 
We design a video-based motion generation module and an audio-based one, which can be plugged into the framework either individually or jointly to drive the video generation.
The predicted motion is used to transform the latent features of StyleGAN for visual animation. 
To compensate for the transformation distortion, we propose a calibration network as well as a domain loss to refine the features.  
Moreover, our framework allows two types of facial editing, \textit{i.e.,} global editing via GAN inversion and intuitive editing based on 3D morphable models. 
Comprehensive experiments show superior video quality, flexible controllability, and editability over state-of-the-art methods.

\end{abstract}



\begin{teaserfigure}
\centering
  \includegraphics[width=\textwidth]{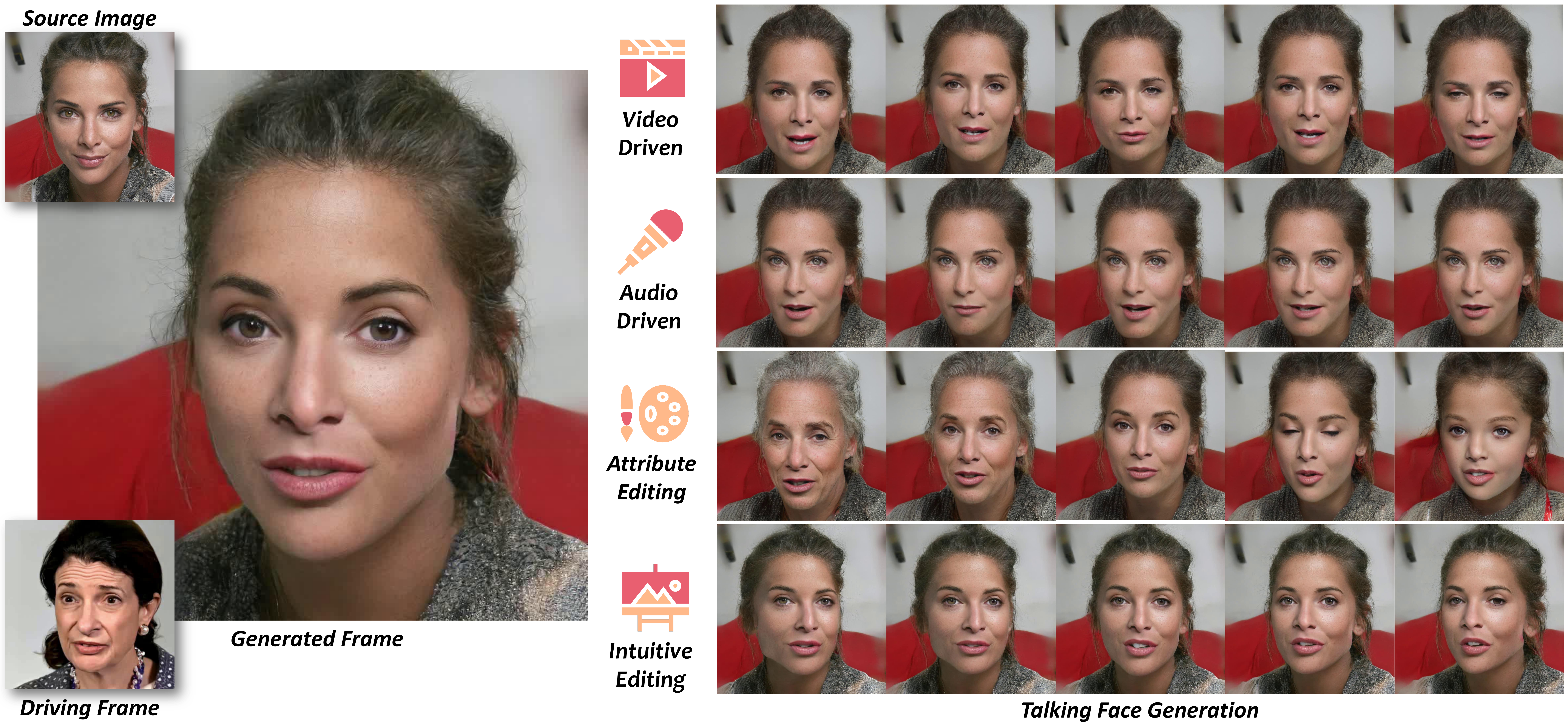}
  \vspace{-4mm}
  \caption{Our unified framework enables high-resolution talking face generation, disentangled control by a driving video or audio, and flexible face editing. 
  Our method elevates the resolution of one-shot talking face generation to 1024$\times$1024 for the first time. 
  The first row shows a synthetic video of video-driven cross-identity reenactment. 
  The second row shows a synthetic video of audio-driven lip movement generation. 
  The third row shows that we can freely edit facial attributes at any timestamp consistently via GAN Inversion during the talking video generation. 
  The fourth row shows that we can perform intuitive face editing based on embedded 3DMM along with talking face generation. 
  }
  \label{fig:teaser}
\end{teaserfigure}
\settopmatter{printacmref=false} 
\renewcommand\footnotetextcopyrightpermission[1]{} 
\settopmatter{printacmref=false,  printccs=false,  printfolios=false}
\maketitle

\newcommand{\xiaodong}[1]{{\color{blue}{[xiaodong: #1]}}}
\newcommand{\yong}[1]{{\color{red}{[yong: #1]}}}

\section{Introduction}


One-shot talking face generation refers to the task of synthesizing a high-quality talking face video from a given portrait image, guided by a driving video or audio segment. 
The synthesized face inherits the identity information from the portrait image, while its pose and expression are transferred from the driving video or generated based on the driving audio. 
Talking face generation has a variety of important applications such as digital human animation, film production, etc.


There are many attempts to drive a static portrait with a video or audio from different perspectives in recent literature. 
A set of methods~\cite{ren2021pirenderer,zhang2021hdtf,doukas2021headgan,wu2021imitating} take the advantage of 3D Morphable Models~(3DMMs), a parametric model that decomposes expression, pose, and identity, to transfer facial motions.
For the audio-driven case, the audio features are always projected to the parameter space of 3DMM~\cite{Zhang_2021_ICCV,wu2021imitating,yi2020audio}.  
Besides, many methods~\cite{song2021everything, makeittalk,lu2021live,zakharov2019few} use facial landmarks as the intermediate representation. 
Some other model-free methods~\cite{face_vid2vid,wiles2018x2face,monkeynet,siarohin2019fomm,wang2021audio2head} remove face prior by using unsupervised key point detection. 

Among those works, subject-dependent methods tend to achieve higher quality results as they require long video footage of the source subject to learn a person-specific model that fits the training video. 
In contrast, subject-agnostic methods aim at obtaining a generic model that is applicable to any source subject, at the cost of synthesis quality due to the limited information about the source. 
One-shot talking face generation is under the subject-agnostic setting, which is much more challenging given the minimal input. 


Recent one-shot talking face generation methods~\cite{zhang2021hdtf,face_vid2vid,ren2021pirenderer} have made notable progress in driving expression and pose, however, they fail to generate high-resolution video frames. 
The video resolution of the common methods still remains at 256$\times$256. Few methods such as~\cite{face_vid2vid} and \cite{zhang2021hdtf} have achieved the resolution of 512$\times$512 by exploiting newly collected high-resolution datasets, \textit{i.e.,} TalkingHead-1KH and HDTF, but they are still bounded by the resolution of the training data. 
More importantly, improving the resolution requires properly designed network architectures and training strategies. 
Adding upsampling layers in a straightforward way into the network usually does not work well.  

We raise an ambitious question: can we further improve the resolution of one-shot talking face to 1024$\times$1024 even though the existing datasets have a lower resolution? 
To achieve this goal, we resort to a powerful pre-trained generative model: StyleGAN~\cite{StyleGAN}. 
StyleGAN has shown impressive results in various applications, \textit{e.g.,} facial attribute editing~\cite{chen2020sofgan}, blind image restoration~\cite{wang2021gfpgan}, portrait stylization~\cite{song2021agilegan}, etc. 
A set of GAN inversion techniques~\cite{stylerig,styleflow,wang2021hfgi,zhu2021barbershop} have achieved high-quality results in face manipulation based on a pre-trained StyleGAN. 
These methods utilize the learned image prior of StyleGAN to facilitate downstream tasks, removing the need of training a large model from scratch. 
The image resolution is retained at 1024$\times$1024 and visual details are also reserved. 
Despite these successes, to the best of our knowledge, there is no existing work that uses a pre-trained StyleGAN for one-shot talking face generation.

In this work, we first investigate the latent style space and the feature space of a pre-trained StyleGAN. 
The style space is also called $\mathcal{W}$ space, which is constructed by mapping a normal distribution to a new distribution via a multi-layer perceptron~(MLP). 
Style codes are used to modify the feature maps of the backbone through AdaIN~\cite{huang2017adain}. 
The style space is extensively explored by GAN inversion methods for face editing. 
The feature space is also called $\mathcal{F}$ space, which is generated by convolution operations in each layer of the backbone. 
It has not been given much attention in the literature:
only a few optimization-based inversion methods~\cite{kang2021out_of_range_inversion,zhu2021barbershop} pay a visit for better reconstruction and attribute editing.  
In a talking-head video, different facial expressions are achieved by deforming different facial regions in different ways. 
Hence, the style space is not an appropriate choice for injecting facial motion information, given that style codes are latent vectors that do not contain accurate spatial information. 
We then systematically study the feature space by applying a set of spatial transformations on the feature map of StyleGAN, including translation, rotation, zooming in, zooming out, shear, occlusion, and warping.  
Interestingly, we discover that the pre-trained model is robust to these operations as it can steadily generate high-quality images accordingly, indicating that the feature space has satisfying spatial properties.  
Fig.~\ref{fig:spatial} shows some synthesized results. 
This investigation reveals that incorporating facial motion information into the feature space is a promising direction for high-quality talking face generation.

%
\begin{table}[t]
\begin{center}
\scalebox{0.83}{
\begin{tabular}{cccccc}
\toprule
 Feature & Resolution & Video & Audio & Intuitive & Attribute \\ 
  &  & Driven & Driven & Editing & Editing \\ \midrule
X2Face~\cite{pix2pix} &  256 & \Checkmark &  &  &  \\
Bi-layer~\cite{zakharov2020bilayer} & 256 & \Checkmark &  &  &  \\
FOMM~\cite{siarohin2019fomm} & 256 & \Checkmark &  &  &  \\
HeadGAN~\cite{doukas2021headgan} & 256 & \Checkmark &  &  \Checkmark & \\  
face-Vid2Vid~\cite{face_vid2vid} & 512 & \Checkmark &  & \Checkmark &  \\
HDTF~\cite{zhang2021hdtf} & 512 &  & \Checkmark &  &  \\
PC-AVS~\cite{pc-talking} & 224 &  & \Checkmark &  \Checkmark &  \\
wav2lip~\cite{prajwal2020wav2lip} & 96 &  & \Checkmark &  &  \\
PIRenderer~\cite{ren2021pirenderer} & 256 & \Checkmark & \Checkmark & \Checkmark &  \\
Ours& \textbf{1024} & \Checkmark & \Checkmark & \Checkmark & \Checkmark \\

\bottomrule
\end{tabular}
}
\end{center}
\caption{Feature comparisons among several one-shot talking face generation methods. Note that `video-driven' refers to the methods which can generate talking faces with only the driving video.  
`Intuitive Editing' refers to editing pose and expression. 
}
\vspace{-2em}
\label{tb:comp_input}
\end{table}


Upon the above observation, we propose a novel unified framework for high-quality one-shot talking face generation based on a pre-trained StyleGAN. 
Our framework enables a set of powerful functionalities, including high-resolution video generation, disentangled control by driving video and audio, and flexible face editing. 
Thanks to the pre-trained StyleGAN, our method can reach the resolution of 1024$\times$1024 without training on new datasets. 
For talking face generation, we exploit the commonly used flow field as the motion descriptor. 
We design a video-based motion generation module to extract motion from video, and an audio-based one to extract motion from audio. 
The predicted flow field is used to spatially warp the latent feature map. However, the warping operation always introduces noticeable artifacts in the final output, especially around the eyes and teeth.  
Hence, we propose a calibration network as well as a domain loss to refine the distorted feature map. 
These two modules can be plugged into the framework either individually or jointly.  
When using both modules, the driving information of pose comes from the video and the driving information of lip movement comes from the audio.  
Furthermore, our framework allows for two types of face editing, \textit{i.e.,} global editing via GAN inversion and intuitive editing based on 3DMM. 
Specifically, given a source portrait, we perform GAN inversion to obtain its style codes that can be used to modify feature maps. This allows us to easily edit global facial attributes via the style codes when generating a talking face video. 
In the video-based motion generation module, we exploit 3DMM parameters to guide the flow field generation, thus intuitive editing can be achieved by modifying the 3DMM parameters. 
Feature comparisons among several related works are presented in Table~\ref{tb:comp_input}. 
Fig.~\ref{fig:teaser} illustrates the functionalities of the proposed framework. 


Our main contributions are as follows:
\begin{itemize}
    \item We propose a unified framework based on a pre-trained StyleGAN for one-shot talking face generation. It enables high-resolution video generation, disentangled control by driving video and audio, and flexible face editing. 
    \item We conduct comprehensive experiments to illustrate the various capabilities of our framework and compare it with many state-of-the-art methods. 
\end{itemize}

\section{Related Work}

\subsection{Talking-head Video Generation.} 

\subsubsection{3D structure-based methods}
Traditionally, 3D faces model priors~(such as 3DMM~\cite{3dmm}) provide a powerful tool for rendering and editing the portrait images by the parameters modulation. For example, DVP~\cite{dvp} modifies the parameters from source and target, then, a network is used to render the shading to video. NS-PVD~\cite{kim2019neural} extends DVP by a novel target-style preserving recurrent GAN. Recent 3D model-based methods~\cite{wpgan,fried2019text,doukas2021headgan,ren2021pirenderer} can also do a good job for subject-agnostic face synthesis. HeadGAN~\cite{doukas2021headgan} pre-processes the 3d mash as input of the network. PIRenderer~\cite{ren2021pirenderer} predicts a flow field for feature warping. Although model-based methods achieve impressive performance, their ability is restricted. Since the 3d models only encode the face region where the realistic associated information~(for example, the hair, teeth, etc) is hard to synthesize. 

\subsubsection{2D-based methods}
Instead of controlling the model parameters, mimicking the motions of another individual by the neural network is also a popular direction. Early works~\cite{wang2018video2video,bansal2018recycle,wu2018reenactgan} learn to map from the source to the target video by image-to-image translation~\cite{pix2pix}. However, these approaches can only work on an individual model of a single identity. 
Later, the meta-learning framework has been explored in fine-tuning models on target identities~\cite{few-vid2vid,zakharov2019few}. These methods use a few target identity samples but fail in complex real-world scenarios. Subject-agnostic approaches~\cite{burkov2020reenactment_latent_pose,anonymous2022latent,siarohin2019fomm,siarohin2021motion,face_vid2vid,monkeynet}, which only need a single image of the target person are the most popular type. For the representative methods, Monkey-Net~\cite{monkeynet} propose a network to transfer the deformation from sparse to dense motion flow. FOMM~\cite{siarohin2019fomm} extends Monkey-Net via the first-order local affine transformations. Then, Face-vid2vid~\cite{face_vid2vid} improves FOMM via a learned 3D unsupervised key-points for free-view talking head generation. Unlike previous methods, \cite{anonymous2022latent} learn to animate the source images via the navigation in latent space. 

\subsubsection{Audio-driven Talking-head Generation.} Another noticeable direction for talking-head generation is audio-driven methods. These methods generate convincing face motions from the audio streams. Early approaches also learn the model for the specific speaker, such as Synthesizing Obama~\cite{obama}, NVP~\cite{NVP}, AudioDVP~\cite{AudioDVP}. Inspired by the recent development of neural rendering, NeRF~\cite{nerf} based talking-head generation has also been proposed~\cite{adnerf}. As for the subject-agnostic methods, reconstruction-based methods can synthesize accurate lips, Speech2Vid~\cite{chung2016syncnet} propose an end-to-end neural network. Then, \cite{zhou2019talking} extends this method via adversarial learning. Next, Wav2Lip~\cite{prajwal2020wav2lip} sync mouth with audio for inpainting-based reconstruction, and PC-AVS~\cite{pc-talking} learn the pose and lip reconstruction by implicit modulation. \cite{wang2021one} utilize the transformer-based network with the pre-trained FOMM method~\cite{siarohin2019fomm}. Recent work also utilize the additional structure information for subject-agnostic methods, \emph{e.g.}, landmarks~\cite{makeittalk, song2021everything}, motion flow fields~\cite{zhang2021hdtf}, and 3d meshes~\cite{lipsync3d}. 
As for the high-quality talking head generation, \cite{zhang2021hdtf} propose a dataset that still suffers from the lip artifacts since this dataset is relatively small.

However, all the previous talking-head methods cannot generate high-resolution videos. On the one hand, it is hard to collect a large amount of high-resolution talking datasets. On the other hand, the network needs to be carefully designed to learn the high-resolution talking faces. Similar to ours, several approaches~\cite{pc-talking, anonymous2022latent} have also utilized the style-based encoder for a talking-head generation. However, all of them focus on the implicit modularization ability of the style convolution rather than the pre-trained GAN prior. In this work, we first demonstrate that the pre-trained StyleGAN network can be used for talking head video generation.

\begin{figure*}[t]
\begin{center}
\centerline{\includegraphics[width=1\linewidth]{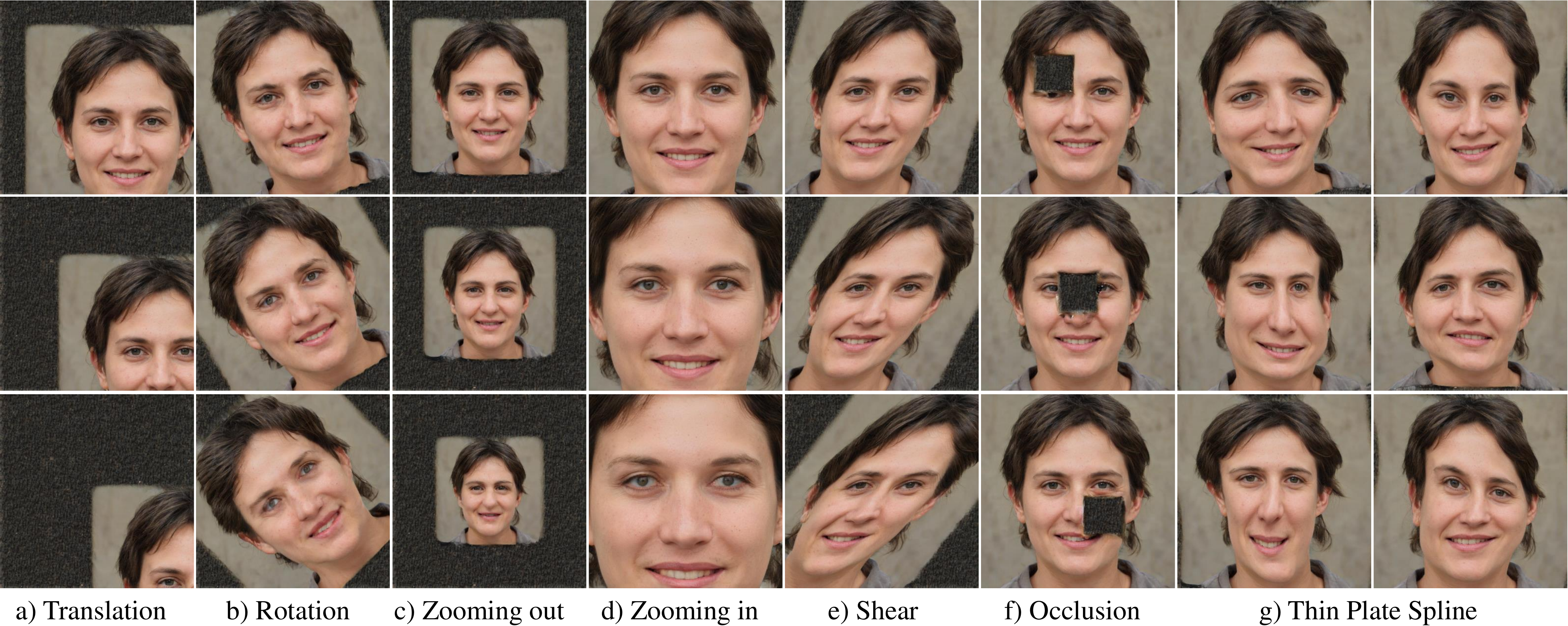}}
\vspace{-2mm}
\caption{
Latent feature space investigation of a pre-trained StyleGAN. 
Different geometric transformations are applied to modify the feature maps.}
\label{fig:spatial}
\end{center}
\end{figure*}

\subsection{Image Editing via Pre-trained StyleGAN}
StyleGAN2~\cite{stylegan2} can generate high-quality face images and draw attention from the community since it can generate high-quality face images and the feature space is highly disentangled.
Thus, StyleGAN editing by GAN Inversion~\cite{zhu2016generative} becomes popular with the rapid evolution of GANs. GAN inversion projects and edits images via the latent space of the pre-trained model. Generally, it can be roughly divided into the optimization-based, encoder-based, and hybrid approaches~\cite{wang2021hfgi}. Optimization-based approaches achieve higher reconstruction quality but need per-image optimization~\cite{abdal2019image2stylegan,image2styleganpp,styleflow,StyleGAN}. 
A more straightforward type is to learn the latent embeddings via additional encoder. pSp~\cite{richardson2021psp} learn the latent space by a UNet-like pixel2style encoder. \cite{wei2021simple, restyle} extend pSp via the multi-stage refinements. 
On the other hand, hybrid methods~\cite{wang2021hfgi,zhu2020domain} are proposed as a combination of the optimization-based and encoder-based methods. 

Besides, we can also classify the GAN inversion by the used latent space. The widely-used inversion space is $\mathcal{W}/\mathcal{W+}$, which is highly-disentangled for the facial attribute editing~\cite{richardson2021psp,wei2021simple,restyle,image2styleganpp,abdal2019image2stylegan,tzaban2022stitch}. There are also some work trying to control the face pose and expression via $\mathcal{W}/\mathcal{W+}$ space. StyleRig~\cite{stylerig} present a rig-like control over a fixed StyleGAN via 3DMM, which can translate the semantic edits on 3D face meshes to the input space of the StyleGAN. However, it fails to create consistent face pose editing and the network cannot generate a novel pose. Several works also try to edit the pre-trained StyleGAN $\mathcal{W}$ space for unconditional video generation~\cite{stylegan_v, tian2021mocogan-hd, stylevideogan} and it is easy to lose the identity. 

On the other hand, the spatial feature space $\mathcal{F}$ is also a promising direction for accurate and local editing. For example, Barbershop~\cite{zhu2021barbershop} uses the segmentation mask and $\mathcal{F}$ space editing for accurate image compositing. GFP-GAN~\cite{wang2021gfpgan} use the $\mathcal{F}$ space for blind face restoration. StyleMapGAN~\cite{kim2021stylemapgan} edit the latent space in spatial dimension for local editing and semantic manipulation. In this work, we give a detailed geometric transformation of the face editing on $\mathcal{F}$ space and propose a method to generate the talking-head video using pre-trained StyleGAN, which preserves the quality and edit-ability of the GAN model and produce time-consistency video content by the conditional signals.

\section{Investigating  Feature Space of StyleGAN}
\label{Spatial Prior}


To allow a pre-trained StyleGAN~\cite{stylegan2} for high-resolution talking-head video generation, one possible direction is StyleGAN based video generation~\cite{tian2021mocogan-hd,stylevideogan}, where they learn to generate videos via discovering an ideal trajectory in $\mathcal{W}^{+}$ latent space.
However, the motion is randomly sampled without any control and the content is corrupted when the current pose differs from the initial one. 
This is because $\mathcal{W}^{+}$ is a highly semantic-condensed space and lacks explicit spatial prior~\cite{wang2021hfgi}.
Moreover, editing in $\mathcal{W}^{+}$ space only allows changing high-level facial attributes, which cannot generate out-of-alignment images~\cite{kang2021out_of_range_inversion} since the StyleGAN is trained on aligned faces.





\begin{figure}[t]
    \centering
    \includegraphics[width=1\linewidth]{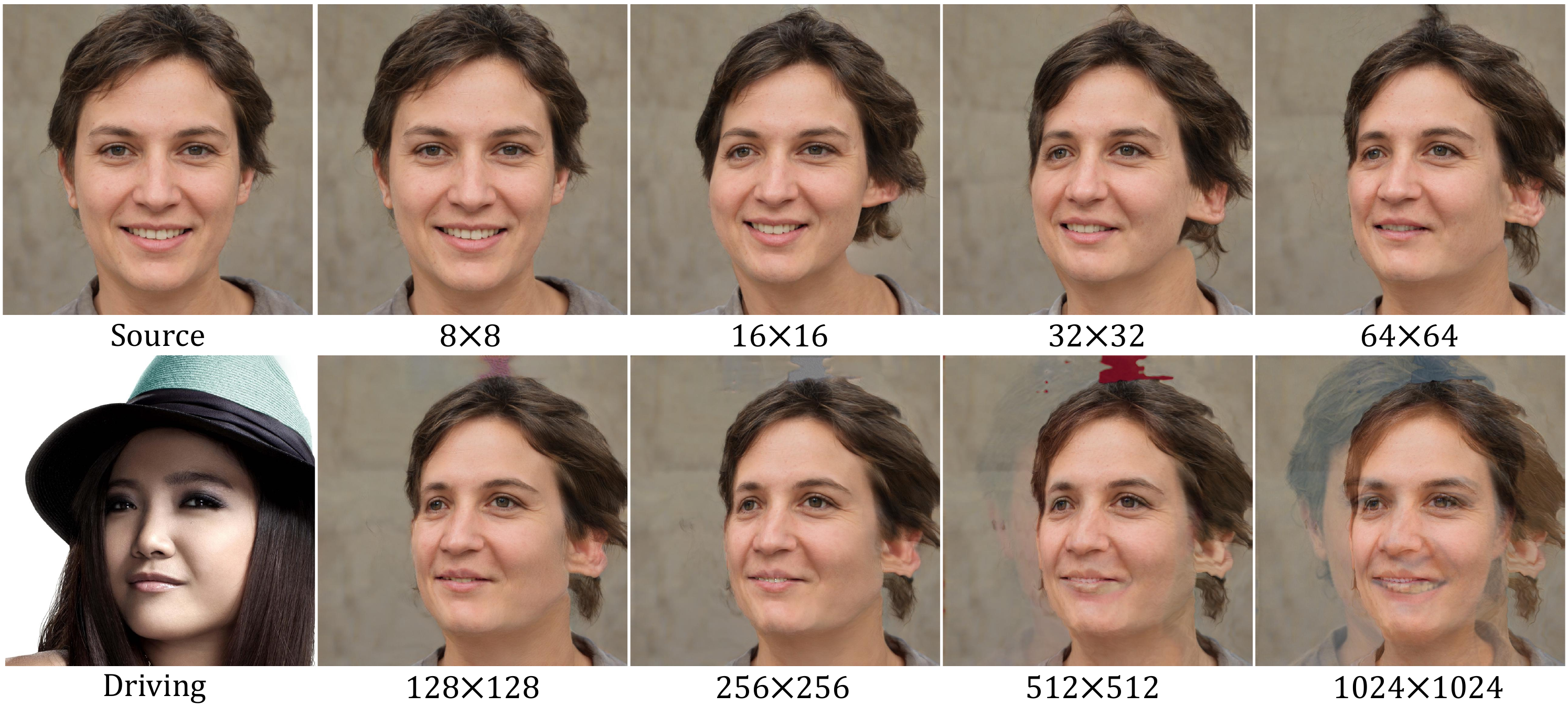}
    \caption{Experiment for which layer to operate on.}
    \vspace{-2em}
    \label{fig:spatial_ablation}
\end{figure}

\if
As a common sense, an image is built by the structure information and the semantic structure, which implies it is hard to directly apply editing operation on them\xiaodong{?}. 
Hence, we need to seek an intermediate state, which maintains the attribute from one-shot source image and can edit the structure continuously by the driven signals.
\fi



\if 
To increase the fidelity and resolution of synthesis, we attempt to leverage the generative facial prior encapsulated in the pre-trained StyleGAN~\cite{stylegan, StyleGAN}.
MoCoGAN-HD~\cite{tian2021mocogan-hd} attempt to render high-resolution videos via discovering an ideal trajectory in $\mathcal{W}^{+}$ space of the pre-trained StyleGAN, in which content and motion are disentangled. Yet the motion is randomly sampled without controlling and the content will be corrupted when the pose differs from the initial situation. It is hard to realize the disentangling of motion and content in $\mathcal{W}^{+}$ space, which is a semantic highly condensed space and lacks the explicitly spatial prior. As common sense, RGB-images possess the explicit spatial prior while keep sparse semantic information, which implies it is hard to directly apply editing operation on them. Hence, we need to seek a intermediate state maintaining both kinds of information between the low-level and high-level states. 
\fi

\begin{figure*}[t]
\begin{center}
\centerline{\includegraphics[width=1\linewidth]{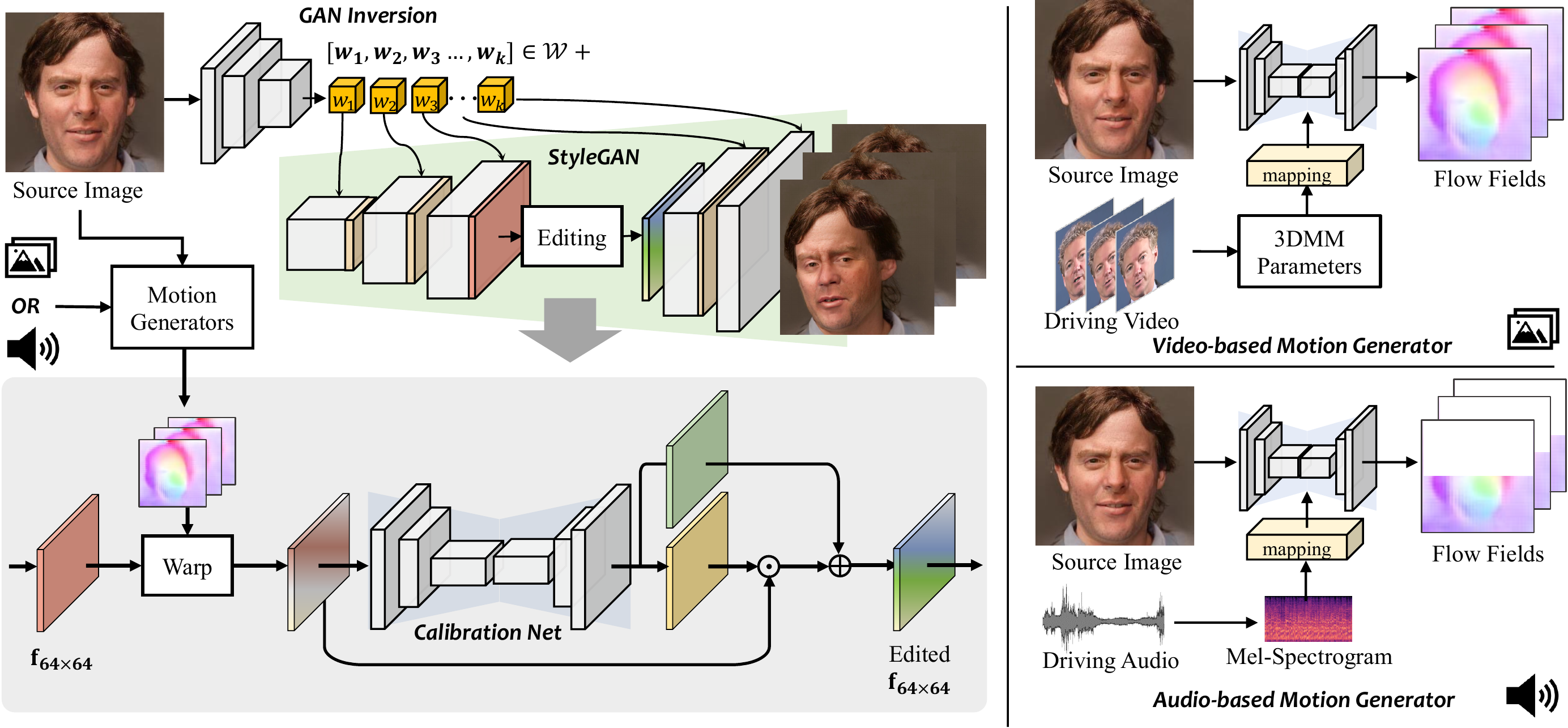}}
\caption{The Pipeline of our unified framework. 
The framework consists of four components, \textit{i.e.,} a pre-trained StyleGAN, a video-driven motion generator, an audio-driven motion generator, and a calibration network. 
Given a source image and a driving video or audio, we can obtain the style codes and feature maps of the source image by the encoder of GAN inversion. 
The video or audio along with the source image are used to predict motion fields by the corresponding motion generator.
The selected feature map is warped by the motion fields, followed by the calibration network for rectifying feature distortions. 
The refined feature map is then fed into the StyleGAN for the final face generation. 
}
\label{fig:pipeline}
\end{center}
\end{figure*}

Thus, image editing in $\mathcal{F}$ feature space~\cite{wang2021hfgi, zhu2021barbershop, kang2021out_of_range_inversion, wang2021gfpgan} draws our close attention. 
Specifically, the latent code $\boldsymbol{f}$ in $\mathcal{F}$ feature space represents a spatial feature map in the generator. 
For StyleGAN~\cite{stylegan2}, we define $\boldsymbol{f}$ as the feature map after a pair of upsampling and convolution layers at a certain scale. 
There are only a few previous methods~\cite{wang2021hfgi, zhu2021barbershop, kang2021out_of_range_inversion, wang2021gfpgan} that edit the spatial features for GAN inversion~\cite{wang2021hfgi,kang2021out_of_range_inversion}, image composition~\cite{zhu2021barbershop}, and blind face enhancement~\cite{wang2021gfpgan}.
These approaches harvest the potential of spatial feature space editing and apply the spatial modulation~(\emph{e.g.} spatial feature transformation~\cite{wang2018sft}) to the features. 
However, it has not been fully investigated whether the feature space of a pre-trained StyleGAN can still be used to generate realistic images after various geometric transformations.

We therefore conduct a detailed experiment to verify the spatial property of StyleGAN features and fully excavate its potential capability. 
We first randomly sample the style latent code $\bm{w}$ in $\mathcal{W}^{+}$ space to generate a random face image with the pre-trained StyleGAN.
At the same time, various spatial features $[\boldsymbol{f}_{4 \times 4}, \boldsymbol{f}_{8 \times 8}, \dots , \boldsymbol{f}_{1024\times 1024}]$ in $\mathcal{F}$ space can be obtained. 
To determine the proper layer for performing spatial transformation, we warp the feature map of each layer individually. 
The results are shown in Fig.~\ref{fig:spatial_ablation}. 
We can observe that warping lower layers cannot accurately control pose and expression while warping higher layers yields ghost shadows on the synthetic image.
Hence, we choose the layer $\boldsymbol{f}_{64 \times 64}$ as a balanced choice. 
Then, to test the spatial property of the pre-trained StyleGAN features, several geometric transformations, including translation, rotation, zoom, shear, and Thin Plate Spline~(TPS~\cite{tps}), \emph{etc.}, are used to manipulate $\boldsymbol{f}_{64 \times 64}$ directly. 
Finally, the transformed image can be generated by the forward pass with the edited feature map as input.

Our experimental results are shown in Fig.~\ref{fig:spatial}. 
Firstly, as shown from a) to e), we apply different affine transformations to $\bm{f}_{64 \times 64}$ with a fixed $\bm{w}$ and feed the edited feature maps into StyleGAN. 
Values are padding with $0$'s for the occlusion. 
We can observe that the generated images share the same identity and appearance with a minor difference. 
This phenomenon demonstrates that the learned convolutional kernels in the pre-trained generator perform in a translation-invariant manner. 
Then, we remove some random patches of the feature map as shown in Fig.~\ref{fig:spatial} f), where the images can still be generated. 
It means that the StyleGAN feature space is also robust to this kind of modification.
Finally, when the complicated deformations, such as TPS operations, are applied to the feature map, the source image is also interpolated to match the randomly sampled target keypoints. 
Overall, either with simple affine transformations or complicated TPS deformations, we observe that the generated images maintain the same geometric changes as the deformations applied in the feature space.

We summarize the strong spatial prior of the intermediate features as follows.
Suppose that image $I$ is generated from feature map $\boldsymbol{f}$ and style code $\boldsymbol{w}$, \emph{i.e.}, $I = G(\boldsymbol{f}, \boldsymbol{w})$, where $G$ is the pre-trained generator. 
For a geometric transformation $T$ in the image space, we have
\begin{equation}
    T(\boldsymbol{I}) \approx G(T'(\boldsymbol{f}), \boldsymbol{w}), 
\end{equation}
where $T'$ is a geometric transformation operator in the feature space, corresponding to $T$. 
$T$'s scale is adjusted according to the relative scale of $\boldsymbol{f}$ to $\boldsymbol{I}$. 
When $T$ is downsampled to the same scale as $T'$, their values are close. 
This spatial property makes it a promising direction to edit the feature space of a pre-trained StyleGAN for talking face generation. 





\section{Methodology}

  

We are interested in the task of controllable talking-head generation. 
Let $I$ be the source image and $\left \{ d_1, d_2, \cdots, d_N \right \}$ be a talking-head video, where $d_i$ is the $i$-th video frame and $N$ is the total number of frames. 
An ideal framework is supposed to generate video $\left \{ y_1, y_2, \cdots, y_N \right \}$ with the same identity as $I$ and the consistent motions derived from $\left \{ d_1, d_2, \cdots, d_N \right \}$.


Inspired by our observation in Sec.~\ref{Spatial Prior}, we propose a unified framework based on the $\mathcal{F}$ space excavation of the pre-trained StyleGAN. 
As shown in Fig.~\ref{fig:pipeline}, our approach contains several steps to achieve this goal. 
Given a single source image, we first use the GAN inversion method~\cite{wang2021hfgi} to get the latent style code and feature maps of the source image. 
Then, to inject the accurate motion guidance, we predict a dense flow field by the motion generator from video~(Sec.~\ref{sec:visual_motion}) or audio~(Sec.~\ref{sec:audio_reenactment}) directly. 
Finally, since the warping operation may introduce artifacts due to the occlusions and error mapping, a calibration network is introduced to renovate the edited spatial feature map~(Sec.~\ref{sec:calibration}). 
%
In the following, we discuss each part in detail.

\subsection{Video-Driven Motion Generator}
\label{sec:visual_motion}
The goal of the video-driven motion generator is to generate dense flows with the driving video and the source image as inputs. 
Then, these flow fields will manipulate the feature map of the pre-trained StyleGAN for talking face generation. 
In this part, we first demonstrate the intermediate motion representation in our settings.  
Then, we give the details of the network structure and the training process for the dense motion field generation.

\paragraph{Motion Representation.}
To achieve accurate and intuitive motion control, semantic medium plays an important role in the generation process. 
Following previous works~\cite{ren2021pirenderer, doukas2021headgan}, we take advantage of the 3DMM~\cite{blanz1999morphable} parameters for motion modeling. 
In 3DMM, the 3D shape $\bm{S}$ of a face can be decoupled as:
\begin{equation}
    \bm{S} = \overline{\bm{S}} + \bm{\alpha} \bm{U}_{id} + \bm{\beta} \bm{U}_{exp},
\end{equation}
where $\overline{\bm{S}}$ is the average shape, $\bm{U}_{id}$ and $\bm{U}_{exp}$ are the orthonormal basis of identity and expression of LSFM morphable model~\cite{3dmm}. 
Coeffcients $\bm{\alpha} \in \mathbb{R}^{80}$ and $\bm{\beta} \in \mathbb{R}^{64}$ describe the person identity and expression, respectively. 
To preserve pose variance, coefficients $\bm{r} \in SO(3)$ and $\bm{t} \in \mathbb{R}^{3}$ denote the head rotation and translation. 
Then, we can model the motion of the driving face with a parameter set $\bm{p} = \left \{ \bm{\beta}, \bm{r}, \bm{t} \right \}$ extracted by an existing 3D face reconstruction model~\cite{deng2019accurate3dmm}.

Due to the inevitable prediction errors between consecutive frames in the same video, the parameters from a single input frame will cause jitter and instability in the finally generated video. 
Hence, we adopt a windowing strategy for better temporal consistency, where the parameters of the neighboring frames are also taken as the descriptor of the center frame to smooth the motion trajectory. 
Thus, the motion coefficient of the $i$-th driving frame is defined as:
\begin{equation}
    \bm{p}_{i} \equiv \bm{p}_{i-k:i+k} \equiv \left \{ \bm{\beta}_{i-k}, \bm{r}_{i-k}, \bm{t}_{i-k}, \dots, \bm{\beta}_{i}, \bm{r}_{i}, \bm{t}_{i}, \dots, \bm{\beta}_{i+k}, \bm{r}_{i+k}, \bm{t}_{i+k} \right \},
\end{equation}
where $k$ is the radius of the window.

\paragraph{Network Structure.}
Our network is built on a U-Net structure that requires the source image and the driving video as inputs, and the outputs are the desired flow fields for feature warping. 
It contains a 5-layer convolutional encoder and a 3-layer convolutional decoder for multi-scale feature extraction.
We use the 3DMM parameters $\bm{p}_{t}$ from the driving frame $d_t$ as the motion representation.
Specifically, these parameters are first mapped to a latent vector via a 3-layer MLP to aggregate the temporal information. 
Then, the motion parameters are injected into each convolutional layer via the adaptive instance normalization~(AdaIN~\cite{huang2017adain}), which is defined as:
\begin{equation}
    AdaIN(\bm{x}_{i}, \bm{p}) = M_{s}(\bm{p})_{i} \frac{\bm{x}_{i} - \mu(\bm{x}_{i})}{\sigma(\bm{x}_{i})} + M_{b}(\bm{p})_{i},
\end{equation}
where $\mu(\cdot)$ and $\sigma(\cdot)$ represent the average and variance operations, respectively. 
$M_{s}$ and $M_{b}$ are used to estimate the adapted mean and bias value according to the target motion. 
Each feature map $\bm{x}_{i}$ in $W$ is first normalized and then scaled and biased using the corresponding scalar components.
Then, the network can be trained by the source image $I$ and the motion condition $\bm{p}_{t}$ as inputs.
Finally, the loss functions will be calculated between the target image $d_t$ and the generated image by the backward warping, which will be discussed later.

\if
After achieving the flow fields $\bm{w}$, the rough results $\hat{\bm{I}}_{w}$ can be calculated by backward warping $\hat{\bm{I}}_{w} = \bm{I} \circ \bm{w}$ where $\circ$ represents the warping procedure.
\fi

\paragraph{Pre-training Strategy.}
Since we only require a low-resolution flow field to drive the spatial feature map of StyleGAN, our motion generator is pre-trained on the widely-used talking-face datasets~(VoxCeleb~\cite{nagrani2017voxceleb}) to generate trustful flow fields. 
Specifically, as the ground truth flow fields are not available, we predict the flow fields $\bm{n}$ using the network, and then the source frame $\bm{I}$ will be used to calculate the warped frame by $\hat{\bm{I}}_{n} = \bm{I} \circ \bm{n}$. 
Then, given the target frame $\bm{I}_t$, we use the perceptual loss~\cite{johnson2016perceptual_loss} to calculate the $\mathcal{L}_{1}$ distance between the activation maps of the pre-trained VGG-19 network~\cite{simonyan2014vgg}.
\begin{equation}
    \mathcal{L}^{v} = \sum_{i} \Vert \phi_{i}(\hat{\bm{I}}_{n}) - \phi_{i}(\bm{I}_{t}) \Vert_{1},
    \label{equ:warp_loss}
\end{equation}
where $\phi_{i}$ denotes the activation map of the $i$-th layer of the VGG-19 network.
Similar to \cite{siarohin2019fomm}, we calculate the perceptual loss on a number of resolutions by applying pyramid down-sampling on $\bm{I}_{t}$ and $\hat{\bm{I}}_{n}$. 
After training, the generated flow field can be used to edit the feature map of StyleGAN.



\subsection{Audio-Driven Motion Generator}
\label{sec:audio_reenactment}

Audio-driven motion transfer is similar to video-driven motion transfer, but this task is more complex since it requires modeling the relationships between audio and face motions.
Some former works attempt to transform audio features into an intermediate medium, \emph{e.g.}, 3D vertex coordinates~\cite{cudeiro20193d_speaking_styles}, facial model parameters~\cite{taylor2017deep_speech_animation}, and 3DMM parameters~\cite{ren2021pirenderer}. 
Then the medium will be converted into facial movements to animate the whole face. 
However, directly predicting the visual semantic parameters from audio information only is a difficult task and the two-stage converting procedure may accumulate more errors.
Consequently, we directly predict the motion from audio features. 
Next, we first introduce the network structure and the motion representation in our settings. And then we give the pre-training strategy for motion generation.

\paragraph{Network Structure.}
The network structure of the audio-driven motion generator is similar to the proposed video-driven motion generator. 
Differently, the driving signal comes from audio. 
Thus, we transform the original audio to Mel-Spectrogram first. 
Then we use an MLP to squeeze the temporal dimension. 
Finally, these features are injected into the network via AdaIN. 




\begin{figure}[ht]
\begin{center}
\centerline{\includegraphics[width=1\linewidth]{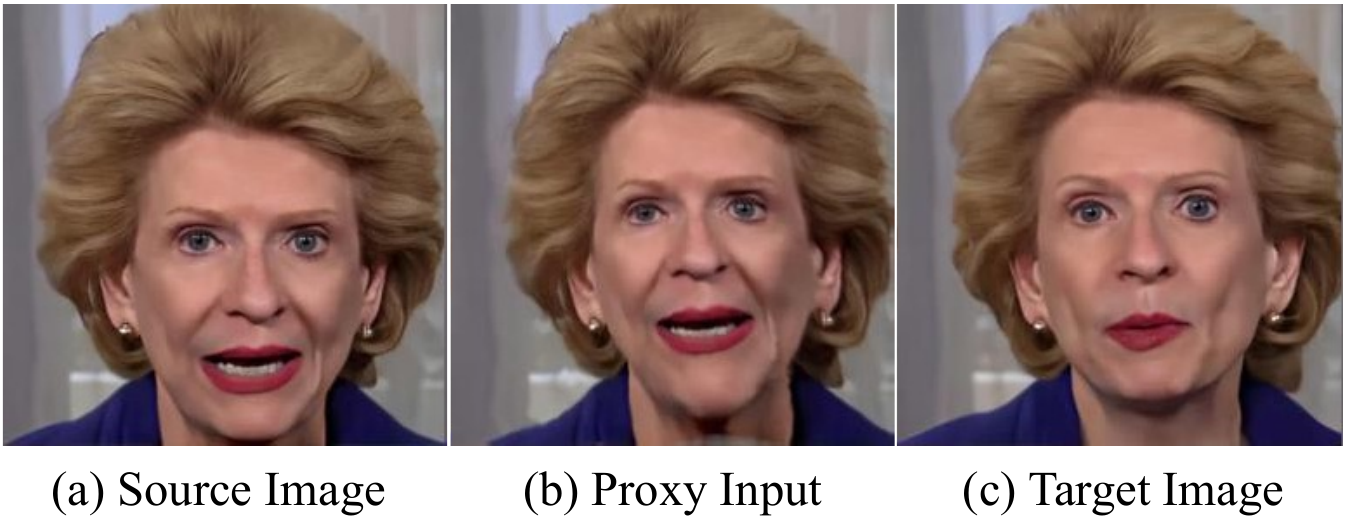}}
\vspace{-1em}
\caption{Paired training data generation for audio-driven motion generator. Thanks to the embedded 3dMM parameters, we use our pre-trained video-based motion generator to create the proxy input, which has the same expression with the source image~(a) and the same head pose with the target image~(c). Then, (b) and the audio from (a) can be used to generate the flow fields with the supervision of (c).
}
\label{fig:proxy}
\vspace{-2em}
\end{center}
\end{figure}

\paragraph{Pre-training Strategy.}
For audio-driven motion generation, we train the generator to predict the flow fields in the lower half face, since audio is closely related to lip movements. 
However, a major challenge, generating a video from audio lacks a paired dataset because the videos with the same pose but different lip shapes are hard to obtain. 
To address this issue, we construct the paired data with the same pose but different expressions under different audio conditions by utilizing the pre-trained video-driven motion generator in Sec.~\ref{sec:visual_motion}. 
Specifically, we generate the proxy input by mixing the 3DMM parameters extracted from the source and driving frame, \textit{i.e.,} the proxy input has the same pose as the driving frame and the same expression as the source frame. 
We illustrate the main process in Fig.~\ref{fig:proxy}, where the head pose of the proxy input is high-aligned with the driving frame. 
By training on the paired dataset, our audio-driven motion generator will focus on the flow generation of expression. 


As for the loss function, similar to our video-driven motion generator, we calculate the loss between the driving image $\bm{I}_t$ and the warped image $\hat{\bm{I}}_{audio}$ using the perceptual loss and the $\mathcal{L}_{1}$ loss. 
Differently, we use a mask strategy to increase the weight of the mouth area. The mask is obtained by calculating the bounding-box of the landmark points around mouth. The loss is defined as:
\begin{equation}
    \begin{aligned}
        \mathcal{L}^a_{t} = &\sum_{i} \Vert M \cdot\phi_{i}( \bm{I}_{t}) - M \cdot\phi_{i}( \hat{\bm{I}}_{audio}) \Vert_{1} + \\ 
        &\lambda^a_{1} \cdot \Vert M \cdot \bm{I}_{t} - M \cdot \hat{\bm{I}}_{audio}) \Vert_{1},
    \end{aligned}
\end{equation}
where $\lambda^a_1$ is the hyper-parameter. 
And in practice, the mask $M$ is in the soft form.

Besides, since the artifacts always happen in the masked region, we design a regularization loss to make sure the consistency of the non-masked region between the proxy input $\hat{\bm{I}}_{visual}$ and the warped image $\hat{\bm{I}}_{audio}$:
\begin{equation}
    \mathcal{L}^a_{reg} = \sum_{i} \Vert(1 - M) \cdot \phi_{i}( \hat{\bm{I}}_{visual}) - (1 - M) \cdot \phi_{i}(\hat{\bm{I}}_{audio}) \Vert_{1},
\end{equation}
Finally, to make the lip movement be more consistent with audio, we employ a lip-sync discriminator $D_{sync}$ that is trained for the synchronization between audio and video by SyncNet~\cite{chung2016syncnet}. 
The synchronization objective can be defined as:


\begin{equation}
    \mathcal{L}^a_{sync} = -\mathbb{E} [ \sum_{t=i-2}^{i+2} \log(D(\hat{\bm{I}}_{audio}, \bm{a})) ],
\end{equation}
where $D_{sync}$ requires $5$ consecutive frames as input.

The total loss can be defined as:
\begin{equation}
    \mathcal{L}^{a} = \mathcal{L}^a_{t} + \lambda^a_{r} \cdot \mathcal{L}^a_{reg} + \lambda^a_{s} \cdot  \mathcal{L}^a_{sync},
\end{equation}
where  $\lambda^a_{r}$ and $\lambda^a_{s}$ are the corresponding weights.

\paragraph{Fully-Controllable Motion Fields.} After training the video-driven and audio-driven motion generator solely, these two generators can be used jointly in a unified framework to control the head motion and the lip movement independently.

\subsection{Feature Calibration and Joint Training}
\begin{figure}[t]
\begin{center}
\centerline{\includegraphics[width=1\linewidth]{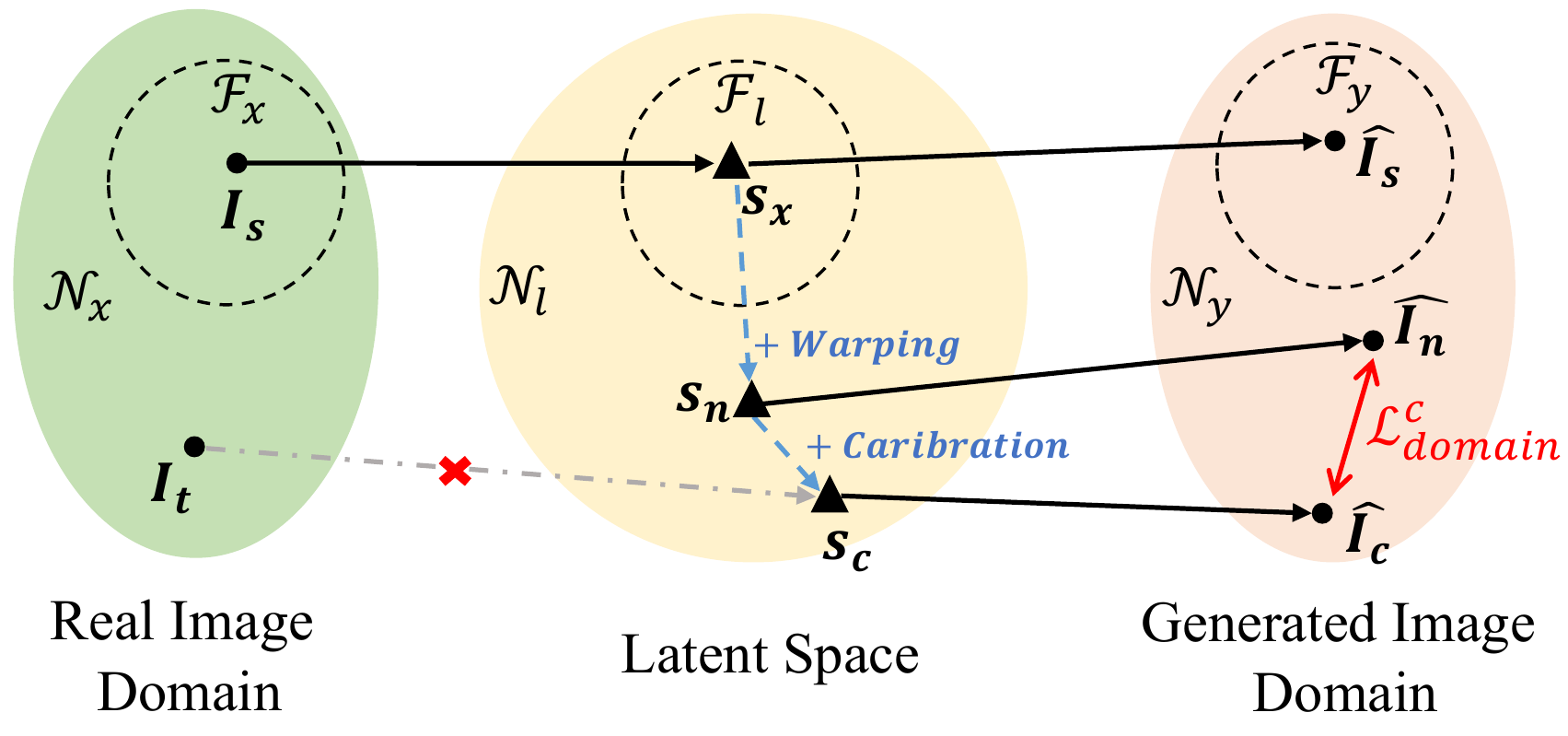}}
\vspace{-1em}
\caption{Illustration of the domain loss.}
\label{fig:regular_loss}
\end{center}
\vspace{-2em}
\end{figure}
\label{sec:calibration}

The video-driven and audio-driven motion generators are pre-trained without considering any information about the pre-trained StyleGAN. 
Though the predicted motion fields can be used to warp the feature map of StyleGAN, it will inevitably introduce artifacts. 
For example, making a closed mouth open through 2D warping cannot fill correct teeth within the mouth. 
To alleviate the feature map distortion, we introduce a calibration network to rectify artifacts in the feature space. 




\paragraph{Calibration Network.}
A calibration network is needed since the warped features still suffer from artifacts. 
As shown in Fig.~\ref{fig:pipeline}, we adopt a U-Net architecture to extract multi-resolution spatial features. It consists of a 4-layer encoder and a 4-layer decoder. 
We feed the warped feature map $\bm{f}_{w}$ as the network's input. 
Then, the multi-scale conventional layers are used to refine the warped features. 
However, due to the high complexity of the intermediate features, 
instead of directly predicting the features, our calibration network performs the spatial
feature transformation (SFT~\cite{wang2018sft}) to the warped features, which is defined as:

\begin{equation}
    \hat{\bm{f}}_{c} = SFT(\bm{f_w}|\bm{\alpha}, \bm{\beta}) = \bm{\alpha} \odot \bm{f_w} + \bm{\beta}
\end{equation}
where $\odot$ denotes element-wise multiplication. 
Then, the final high-quality and high-resolution result can be achieved as $\hat{I} = G(\hat{\bm{f}}_{c}, \bm{w})$.

\begin{figure*}[t]
\begin{center}
\centerline{\includegraphics[width=1\linewidth]{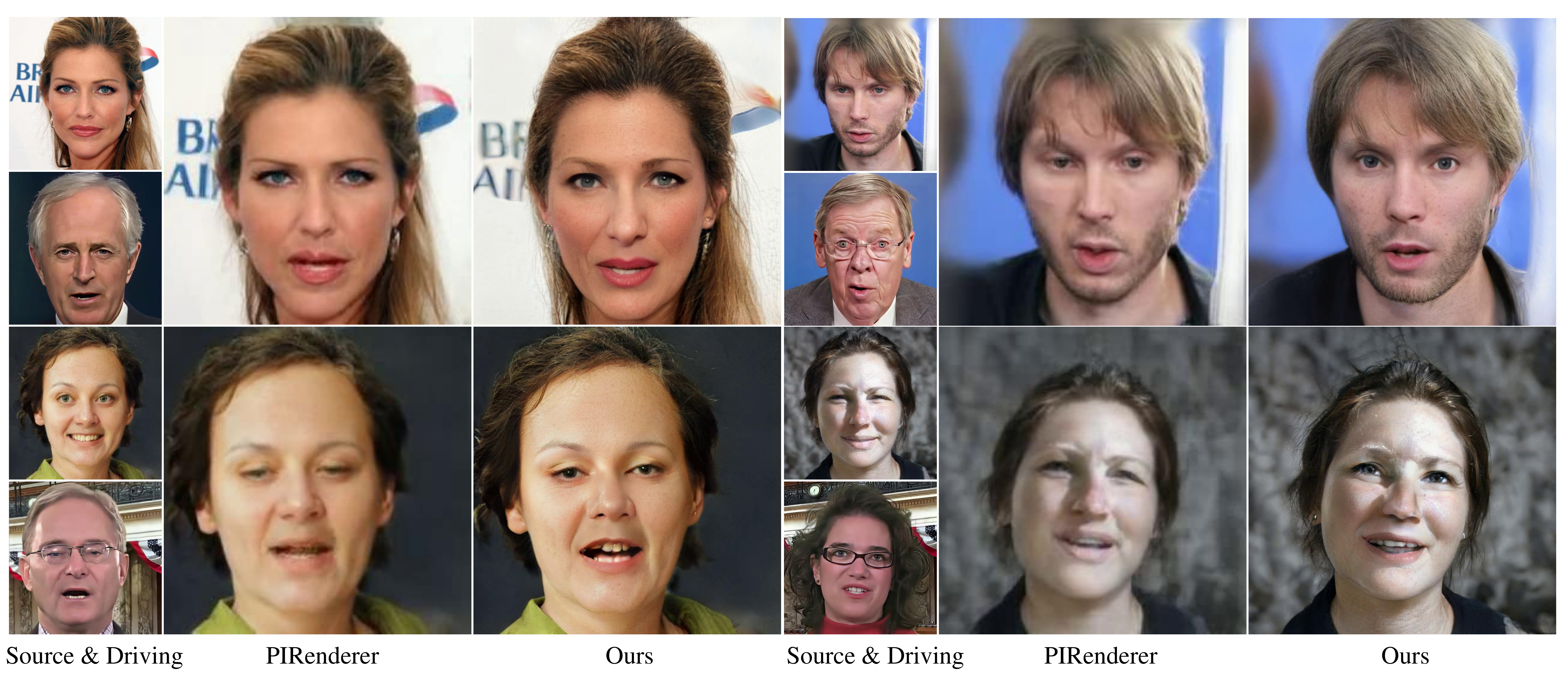}}
\vspace{-1em}
\caption{High-resolution talking face generation. 
Top row: a real face is driven by a real face. 
Bottom row: a synthetic face is driven by a real face. 
Real faces are from HDTF~\cite{zhang2021hdtf}. 
Synthetic faces are sampled from StyleGAN. 
The resolution of images generated by PIRenderer is 256$\times$256 while that of ours is 1024$\times$1024. 
}
\label{fig:resolution}
\end{center}
\vspace{-2em}
\end{figure*}

\paragraph{Overall End-to-end Training.} 
Directly applying the introduced calibration network is easy to encounter blur results~(as shown in Fig.~\ref{fig:loss_ablation}) since the quality of the frames in the video dataset is much lower than the high-resolution face dataset for training StyleGAN.
Furthermore, inevitable detail lost of identity, attribute, texture, and background raised by the GAN inversion method will enlarge the gap between the generated images and the real images, which will further mislead the direction of the optimization. 

Thus, we joint train the whole network and design loss functions to solve the above problem. 
We first design a domain loss to restrict the differences between the reconstructed image of the warped feature map and that of the calibrated feature map in the generated image domain. 
As shown in Fig.~\ref{fig:regular_loss}, given a natural source image $\bm{I}_s$ in the aligned StyleGAN space $\mathcal{F}_x$, the GAN inversion method can invert and reconstruct the image in the latent space and the generated image domain, respectively. 
Differently, for the target image $\bm{I}_t$ which is out of the aligned domain, GAN inversion is hard to be applied.
Thus, to obtain the desired latent space $\bm{s}_c$, the proposed method utilizes the flow fields to edit the images in the latent space. 
After editing, the warped feature $\bm{s}_n$ may not be in the aligned StyleGAN latent space anymore but it can still generate a high-quality image $\hat{\bm{I}}_n$ by forwarding pass as we have discussed in Sec.~\ref{Spatial Prior}. 
Unfortunately, the warping artifacts may occur because of the low quality of the flow fields. 
Thus, we propose the calibration network to further edit the feature map as introduced previously.
However, the results $\hat{\bm{I}}_c$ become blurry due to the feature shift. 
To preserve both advantages of $\hat{\bm{I}}_n$ and $\hat{\bm{I}}_c$, the domain loss is defined to measure their difference. 
Further, we take a masking strategy to enhance the weight of different areas. 
The calibration mask $M$ is comprised of the bounding boxes of the eyes and mouth because the artifacts often occur around them.
Thus, the domain loss is:
\begin{equation}
    \mathcal{L}^c_{domain} = \sum_{i} \Vert (1 - M) \cdot\phi_{i}( \hat{\bm{I}}_{n}) - (1 - M) \cdot \phi_{i}(\hat{\bm{I}}_{c}) \Vert_{1}. 
\end{equation}


\if
In our pipeline, the editing image is achieved via warping the feature latent of source inversion image.
If we directly minimize the distance between editing results and the target images in real image domain (blue line), which is equivalent to minimize both the distance between target inversion images and real target images and the distance between editing results and target inversion images. 
As shown in Fig.~\ref{fig:loss_ablation}, this process will compel calibration network to raise the identity similarity and to cause blurring, which is out of our goal, i.e., calibrating the warping artifacts instead of bridging the gap between two domains.

Supposing the GAN inversion method can be applied on non-aligned images, then the target inversion image is our ultimate goal and the calibration procedure is meant to amend the editing position to the target inversion position.
There exists a problem that the StyleGAN is trained on landmark aligned face images~\cite{bulat2017face_alignment} for performance and the GAN inversion procedure requires the input image to be aligned either. 
As a result, the aligned target inversion images is discontinuous in temporal dimension which loses the temporal information of $\bm{p}$ and will seriously affect the video generation results. 

It is a worthy to mention that the intermediate editing results have been close to the inversion target in the low-frequency part, e.g., the resolution and texture. The warping results can be out of aligned to maintain the temporal information despite the defects of facial features. 

Hence, we decouple the objective as target loss part for artifacts-free purpose and regularize loss part for high fidelity results. 
\fi

Besides, for eliminating the artifacts of local facial features, the driving image $\bm{I}_{t}$ provides the most accurate high-frequency information. 
Hence, we calculate the $\mathcal{L}_{1}$ loss and the perceptual loss with the ground truth, which is weighted on the masked region:
\begin{equation}
    \begin{aligned}
        \mathcal{L}^{c}_t = &\sum_{i} \Vert M \cdot\phi_{i}( \bm{I}_{t}) - M \cdot\phi_{i}( \hat{\bm{I}}_{c}) \Vert_{1} + \\ & \lambda^c_1 \cdot \Vert M \cdot \bm{I}_{t} - M \cdot \hat{\bm{I}}_{c} \Vert_{1},
    \end{aligned}
\end{equation}
where $\lambda^c_1$ is the weight of the $\mathcal{L}_{1}$ loss.


Finally, to maintain the high fidelity of face generation, we also impose adversarial loss. 
Note that we freeze the parameters of the discriminator since the low-quality video frames may decline its performance. The adversarial loss can be defined as:
\begin{equation}
    \mathcal{L}^c_{adv} = -\mathbb{E} [\log(D(\hat{\bm{I}}_{c}))],
\end{equation}
where $D$ is a well-trained discriminator of StyleGAN2.

The framework is trained in an end-to-end manner together with the loss of the corresponding motion generators. 
Here, we calculate the perceptual loss between the intermediate results from the motion generator and the ground truth, which is the same as~Eq.~\ref{equ:warp_loss}. 
The weight of other components~(the StyleGAN generator and the inversion encoder) are frozen. 


\begin{figure}[t]
\begin{center}
\centerline{\includegraphics[width=1\linewidth]{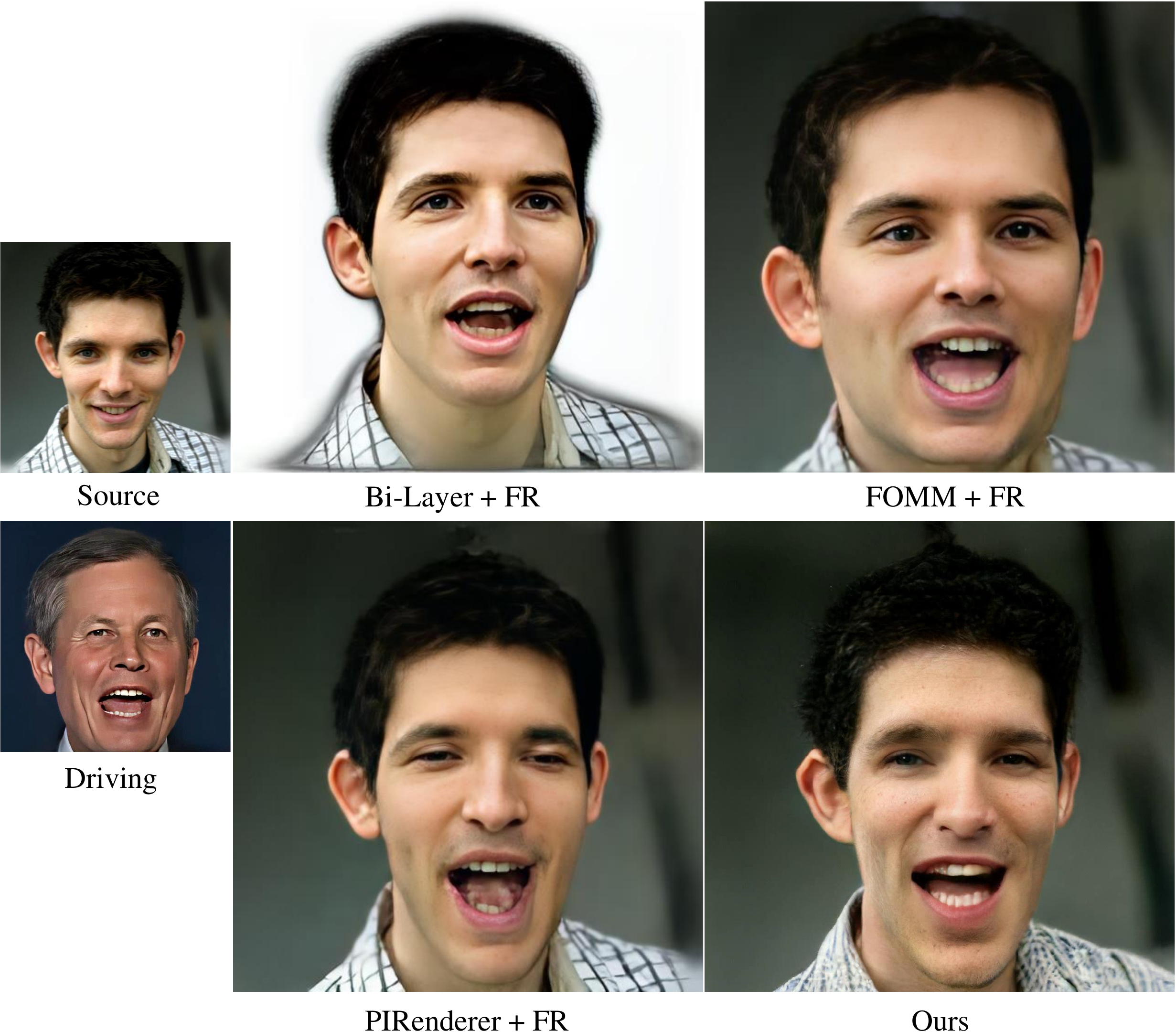}}
\vspace{-1em}
\caption{Comparisons with enhanced methods. 
We use a recent face restoration (FR) method to enhance the visual quality of these competing methods, \textit{i.e.,} GFP-GAN~\cite{wang2021gfpgan}. 
}
\label{fig:sr_ablation}
\end{center}
\vspace{-2em}
\end{figure}

In summary, the overall loss is a weighted summation as follows:
\begin{equation}
    \mathcal{L}^{c} =  \mathcal{L}^c_{t} + \lambda^c_{d} \cdot \mathcal{L}^c_{domain} + \lambda^c_{adv} \cdot  \mathcal{L}^c_{adv} + \beta^{v} \cdot \mathcal{L}^v + \beta^a \cdot  \mathcal{L}^a ,
\end{equation}
where $\lambda^c_{d}$, $\lambda^c_{adv}$, $\beta^v$ and $\beta^a$ are the corresponding weights.

\section{Experiments}

\subsection{Settings}

\paragraph{Datasets.} 
We train the two motion generators on the VoxCeleb dataset~\cite{nagrani2017voxceleb} which consists of over 100K videos of $1,251$ subjects. 
Following~\cite{siarohin2019fomm}, we preprocess the data by cropping faces from the videos and then resizing them to 256$\times$256. 
Faces are not aligned and can move freely within a fixed bounding box. 
We joint train the whole framework on the HDTF dataset~\cite{zhang2021hdtf} which consists of 362 videos of over 300 subjects.
The resolution of original videos is $720P$ or $1080P$, which is higher than that of VoxCeleb. 
The videos are cropped in the same manner as processing VoxCeleb and then resized to  512$\times$512. 
HDTF is split into non-overlapping training and test sets.
The test set contains 20 videos with around 10K frames. 
For cross-identity motion transfer evaluation, we also selected 1,000 high-resolution images from the CelebA-HQ dataset~\cite{karras2017progressive_gan}.

\paragraph{Implementation Details.} 
We train the two motion generators and the calibration network in two stages. 
In the first stage, we pre-train the video-based motion generator on VoxCeleb for 200K iterations. 
Then, we formulate training pairs for the audio-based motion generator by using the predicted motion as the pseudo label. We pre-train the audio-based generator with synthesized audio-motion pairs for 200K iterations. The trade-off hyper-parameters are set to $\lambda^a_{1} = 10$ , $\lambda^a_{r} = 0.1$ and $\lambda^a_{s} = 1$. 
The optimizer for both pretraining processes is ADAM~\cite{kingma2014adam} with an initial learning rate of $10^{-4}$. The batch size is set to 20 for all experiments.

As the motion from the pre-trained generators cannot be seamlessly applied to feature maps of StyleGAN, we need to finetune them along with the calibration network. 
Hence, in the second stage, we first jointly optimize the calibration network and the video-based motion generator in an end-to-end manner on HDTF for 20K iterations. 
The hyper-parameters are set to $\lambda^c_{1} = 10$, $\lambda^c_{d} = 0.01$, $\lambda^c_{adv} = 0.1$, $\beta^v = 0.01$, and $\beta^a = 0$. 
The learning rates are set to $10^{-4}$ and $2 \times 10^{-5}$ for them, respectively. 
Then, we fix the video-based motion generator and jointly optimize the calibration network and the audio-based motion generator for 20K iterations.
The hyper-parameters are set to $\lambda^c_{1} = 10$, $\lambda^c_{d} = 0.01$, $\lambda^c_{adv} = 0.1$, $\beta^v = 0$, and $\beta^a = 0.01$. 
The learning rates are set as the same as the above optimization. 

\begin{figure*}[t]
\begin{center}
\centerline{\includegraphics[width=1\linewidth]{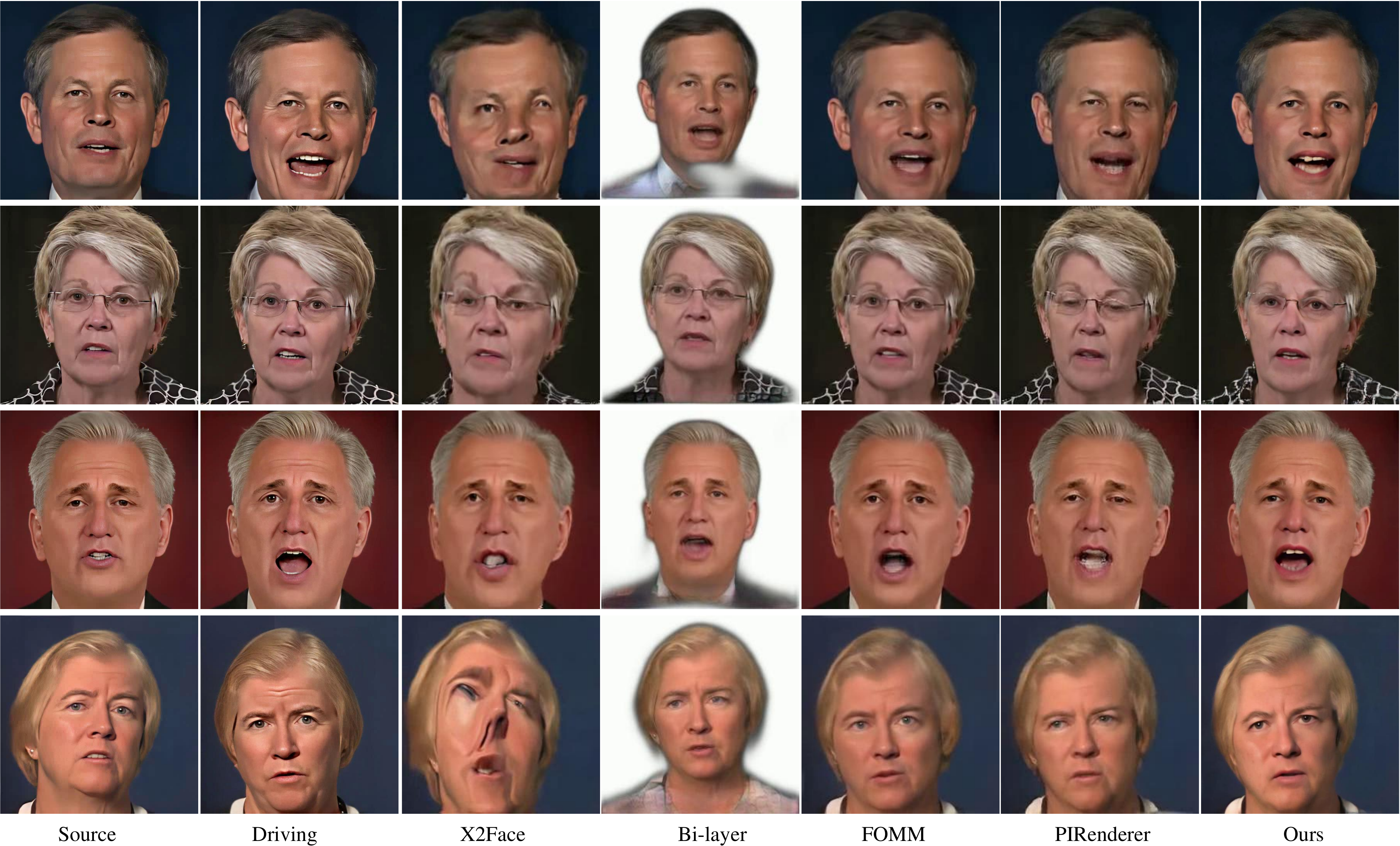}}
\vspace{-1em}
\caption{Qualitative comparisons with state-of-the-art methods on the task of same-identity reenactment.}
\label{fig:same_id}
\end{center}
\vspace{-1em}
\end{figure*}
\begin{table*}[ht]
\begin{center}
\begin{tabular}{c|ccccccc|cccc}
\toprule
& \multicolumn{7}{c|}{\textbf{Same-Identity Reenactment}}  & \multicolumn{4}{c}{\textbf{Cross-Identity Reenactment}} \\ 
\cmidrule{2-12}
& FID $\downarrow$ & LPIPS $\downarrow$  & PSNR $\uparrow$ & SSIM $\uparrow$  & CSIM $\uparrow$ & AED $\downarrow$ & APD $\downarrow$ & FID $\downarrow$ & CSIM $\uparrow$ & AED $\downarrow$ & APD $\downarrow$ \\ 
\midrule


X2Face~\cite{wiles2018x2face}  & $44.32$ & $0.2687$ & $31.09$ & $0.5926$ & $0.6965$ & $0.1680$ & $0.03719$  & $128.19$ & $0.4449$ & $0.3415$ & $0.05156$ \\ 

Bi-layer~\cite{zakharov2020bilayer}  & $118.46$ & $0.5758$ & $28.23$ & $0.2906$ & $0.3033$ & $0.1219$ & $0.01322$  & $189.64$ & $0.2252$ & $0.2654$ & $\bm{0.02054}$  \\ 

FOMM~\cite{siarohin2019fomm}  & $29.17$ & $0.2036$ & $31.12$ & $\bm{0.6353}$ & $\bm{0.8121}$ & $\bm{0.0946}$ & $\bm{0.00914}$  & $108.93$ & $0.4517$ & $0.2692$ & $0.02576$  \\ 

PIRenderer~\cite{ren2021pirenderer}  & $27.14$ & $0.2252$ & $30.96$ & $0.6028$ & $0.7797$ & $0.1073$ & $0.01459$  & $108.56$ & $0.4812$ & $\bm{0.2554}$ & $0.02962$  \\ 

Ours & $\bm{18.02}$ & $\bm{0.1729}$ & $\bm{31.21}$ & $0.6019$ & $0.7475$ & $0.1151$ & $0.01664$  & $\bm{91.28}$ & $\bm{0.4890}$ & $0.2630$ & $0.03484$  \\ 

\bottomrule
\end{tabular}
\end{center}
\caption{Quantitative comparisons with state-of-the-art methods on talking-face motion transfer. 
}

\label{tb:criterion}
\end{table*}

\begin{figure*}[t]
\begin{center}
\centerline{\includegraphics[width=1\linewidth]{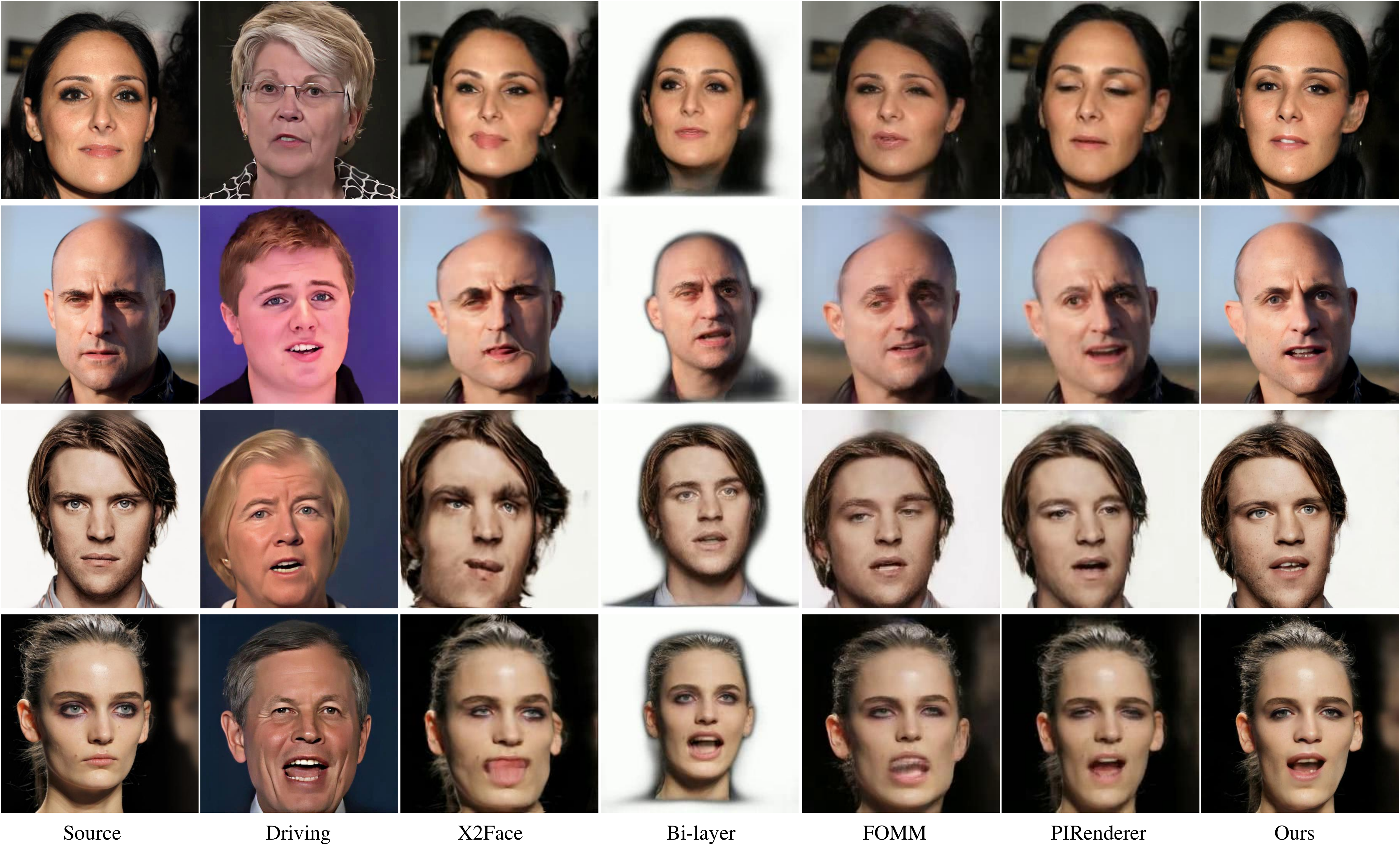}}
\vspace{-1em}
\caption{Qualitative comparisons with state-of-the-art methods on the task of cross-identity reenactment.}
\label{fig:cross_id}
\end{center}
\vspace{-1em}
\end{figure*}

During inference, the two motion generators can be used individually or jointly. 
When using both of them, the video-based motion generator controls the head pose while the audio-based motion generator controls the lip movement. 



The GAN inversion is used to get the spatial feature maps in our framework. 
Almost all existing GAN inversion techniques can be exploited. 
Optimization techniques can achieve more accurate reconstruction results, but they are not efficient. 
While learning-based techniques are much faster, they encounter lower reconstruction quality. 
Considering the efficiency, we exploit a state-of-the-art learning-based inversion method~\cite{wang2021hfgi} during training. 
During inference, we first use~\cite{wang2021hfgi} to obtain the style codes and feature maps. 
For motion transfer tasks, we further exploit an optimization-based inversion method~\cite{zhu2021barbershop} to optimize latent feature maps for more accurate reconstruction.
For editing tasks, we directly use the style codes and feature maps from~\cite{wang2021hfgi}.





\paragraph{Evaluation Metrics.} We exploit a set of metrics to evaluate image quality and motion transfer quality. 
For image quality, Learned Perceptual Image Patch Similarity (LPIPS)~\cite{zhang2018lpips}, Peak signal-to-noise ratio~(PSNR) are utilized as metrics to measure the reconstruction quality. Structural Similarity~(SSIM) is utilized to measure the structural similarity between patches of the input images.
Frechet Inception Distance (FID)~\cite{heusel2017fid} is utilized to measure the realism of the synthesized results.
To measure identity preservation, we compute the cosine similarity~(CSIM) of identity embedding between the source images and the generated videos extracted from ArcFace~\cite{deng2019arcface}.
For motion transfer quality, following~\cite{ren2021pirenderer}, Average Expression Distance~(AED), and Average Pose Distance~(APD) are used to compute the differences between generated images and target images in terms of 3DMM expression and pose, respectively.

\begin{figure*}[t]
\begin{center}
\centerline{\includegraphics[width=1\linewidth]{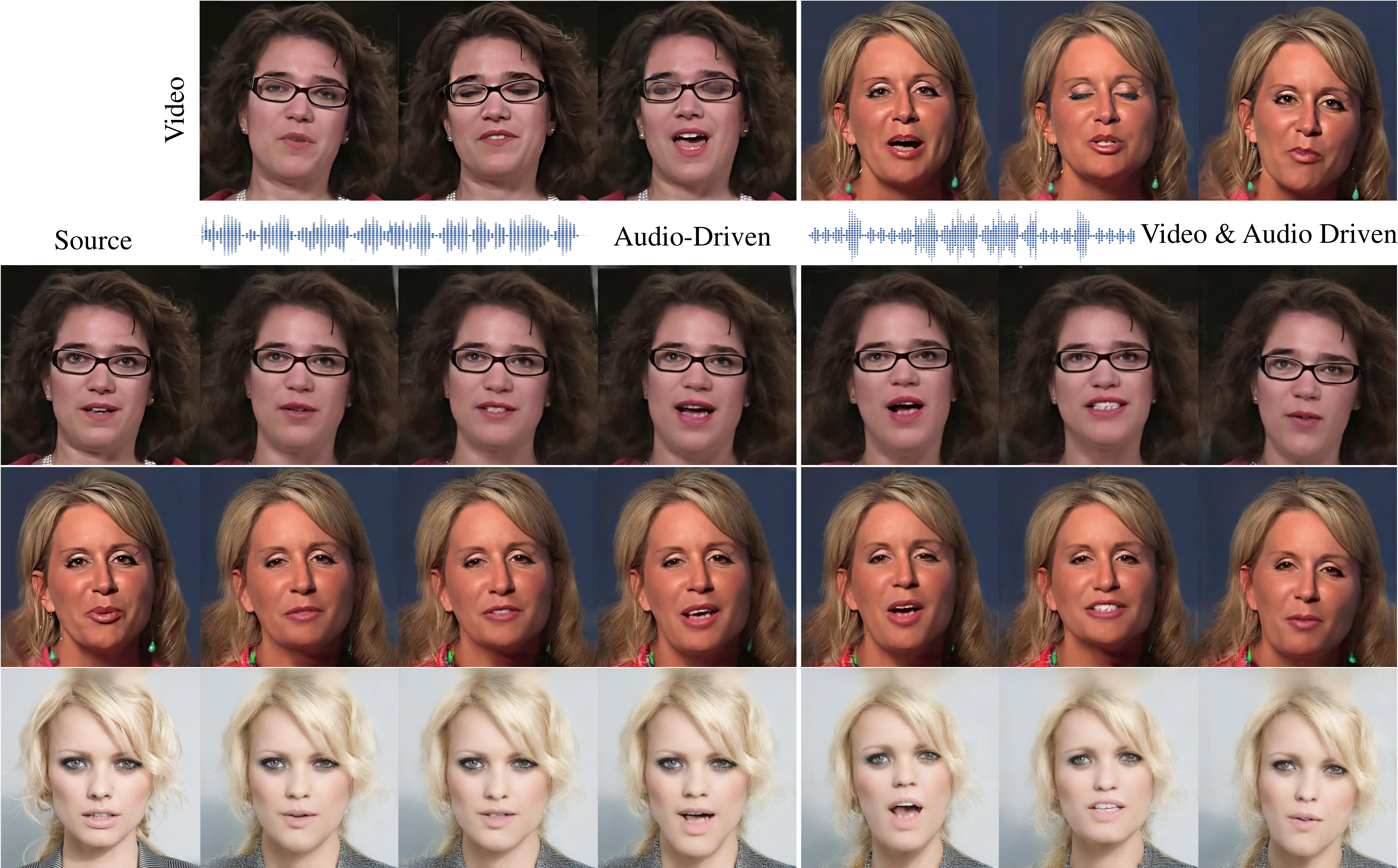}}
\caption{Qualitative results of audio-driven talking face generation. Row 1: videos that provide audios. 
Column 1: source images. 
Column 2-4: audio-driven lip movement generation. 
Column 5-7: video-and-audio jointly driven talking face generation. 
The video controls pose while the audio controls lip movements. 
}
\label{fig:audio}
\end{center}
\end{figure*}

\begin{figure*}[htbp]
    \centering
    \subfigure{
        \includegraphics[width=1\linewidth]{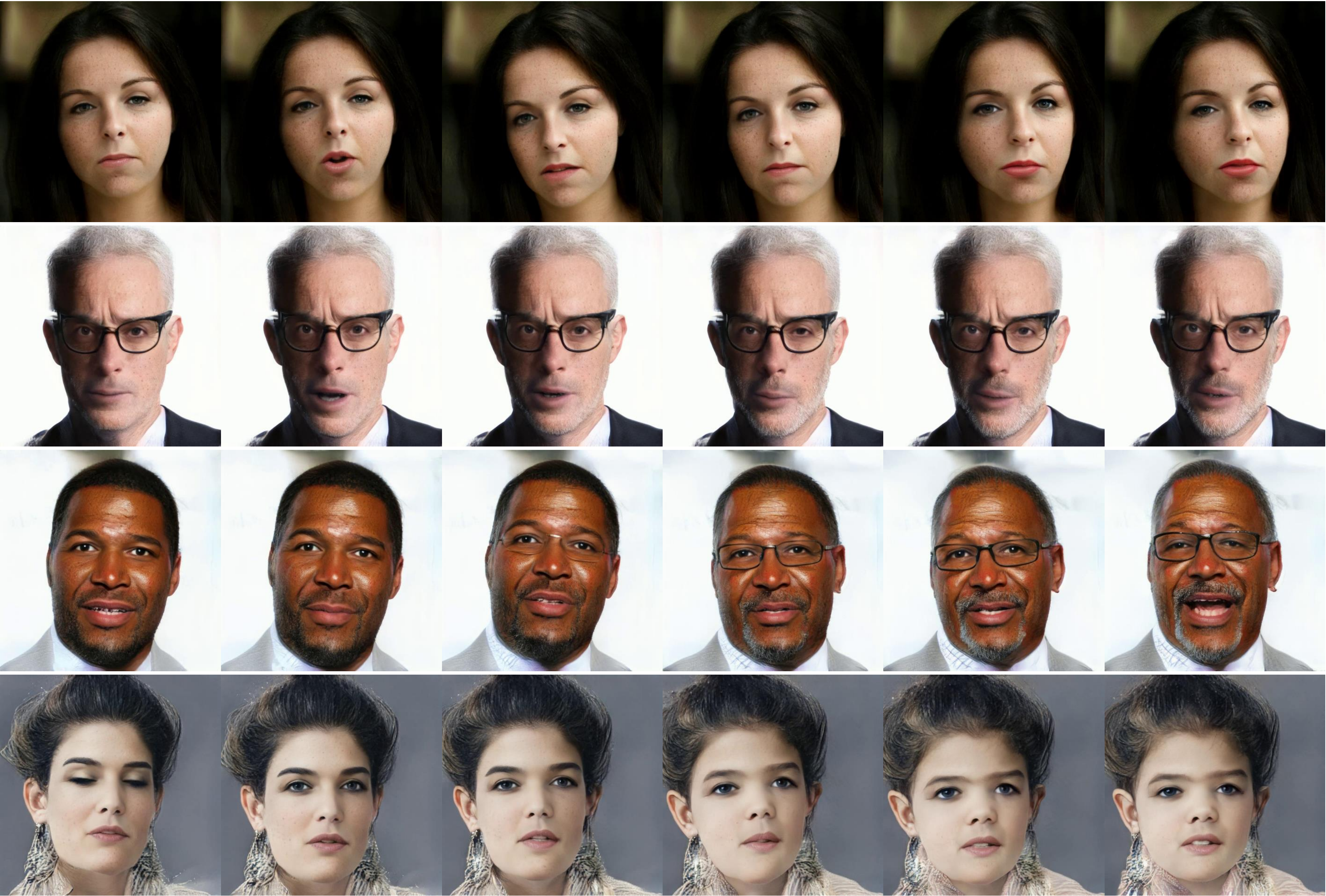}
    }
    \caption{Global attribute editing via GAN inversion. 
    The attribute is gradually modified in each generated talking video. From top to bottom, the edited attributes are adding makeup, adding beard, increasing age, and decreasing age. 
    }
    \label{fig:editing_1}
\end{figure*}


\begin{figure*}[t]
\begin{center}
\centerline{\includegraphics[width=1\linewidth]{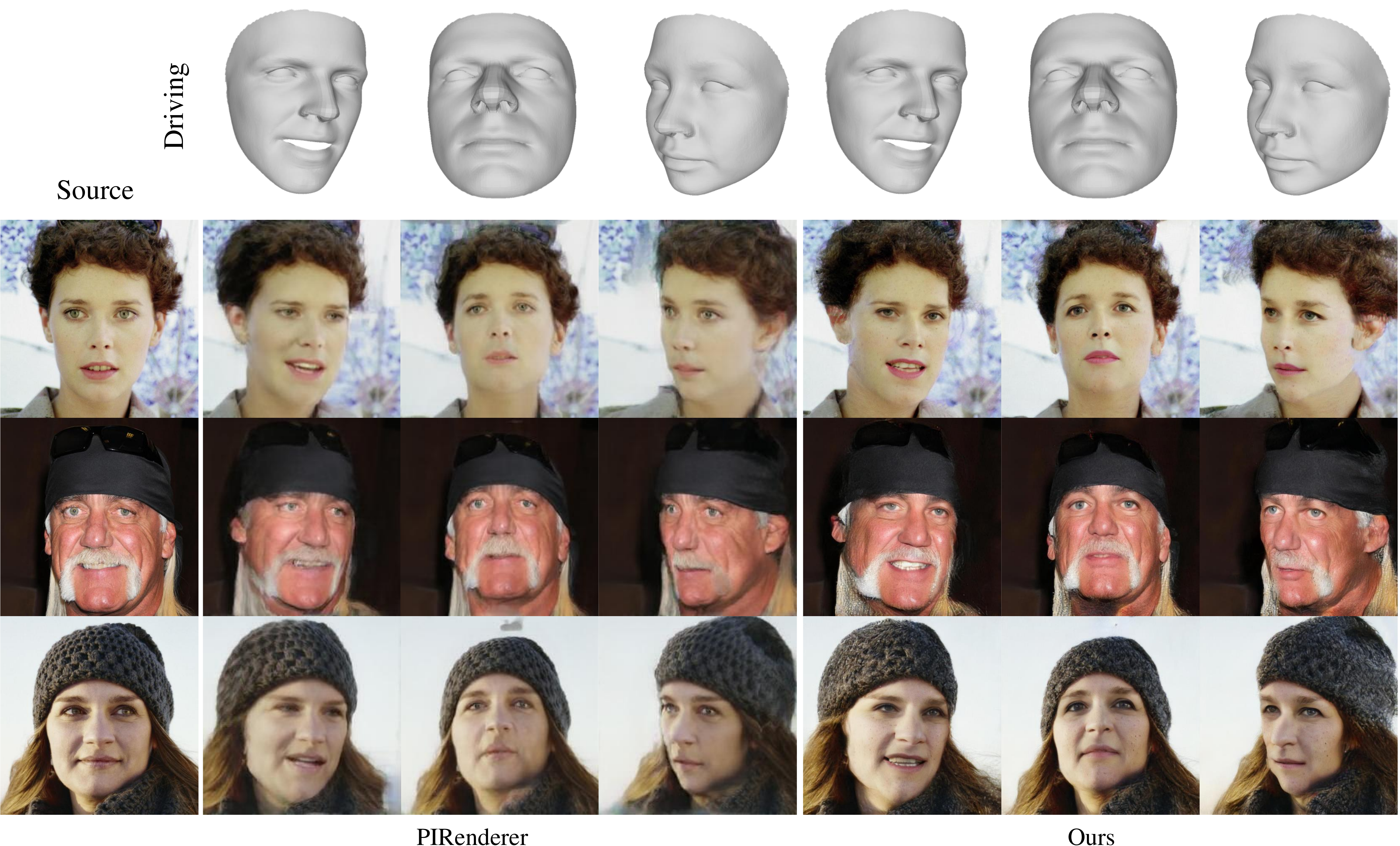}}

\caption{The qualitative results of intuitive face editing. Row 1: the driving 3DMM parameters. Column 1: source images. Column 2-4: synthetic images by PIRenderer. Column 5-7: our synthetic images.}
\label{fig:intuitive_control}
\end{center}

\end{figure*}

\begin{figure}[ht]
\begin{center}
\centerline{\includegraphics[width=1\linewidth]{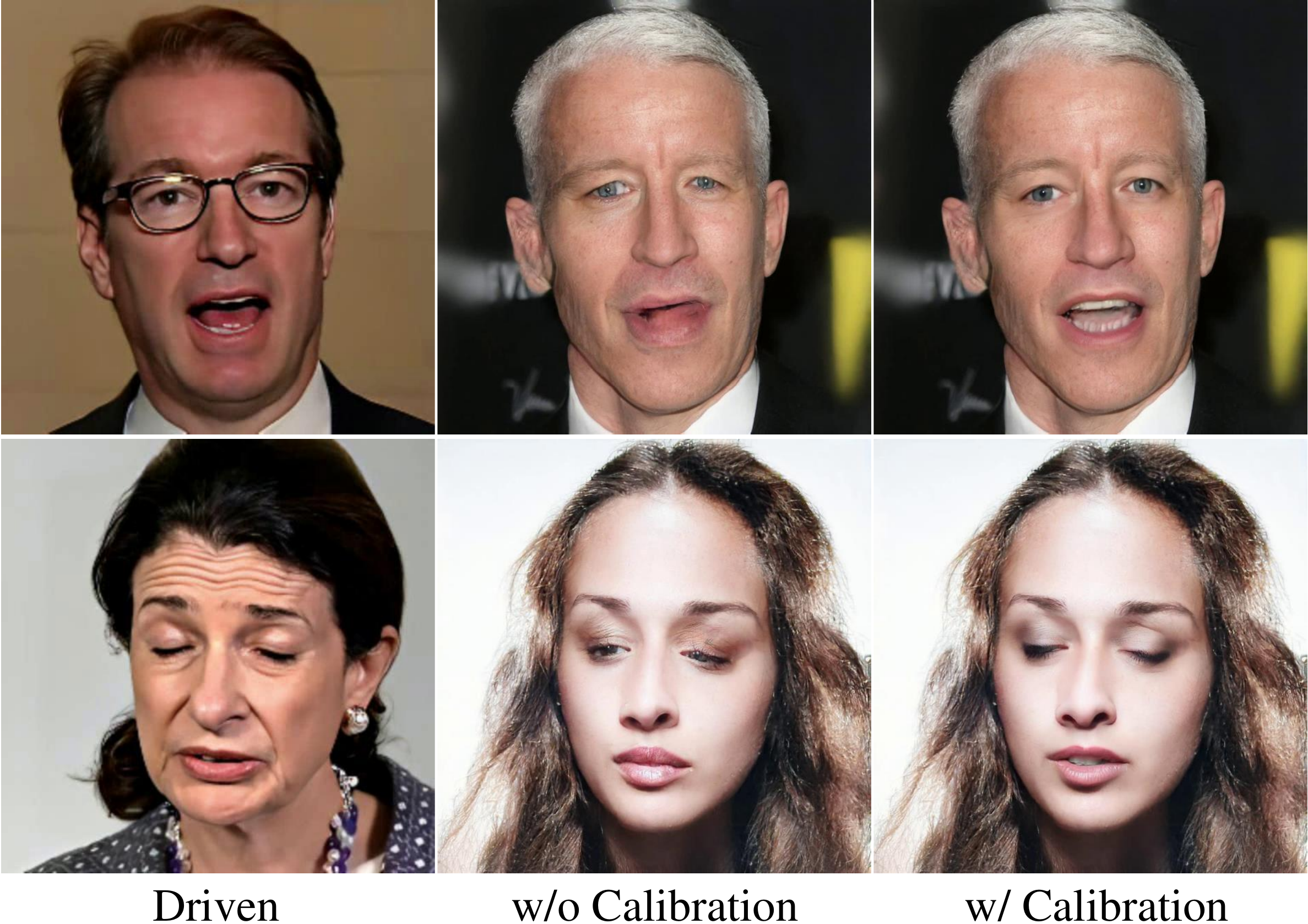}}
\vspace{-0.5em}
\caption{Ablation study of the calibration network. }
\label{fig:calibration_ablation}
\end{center}
\end{figure}
\begin{figure}[ht]
\begin{center}
\centerline{\includegraphics[width=1\linewidth]{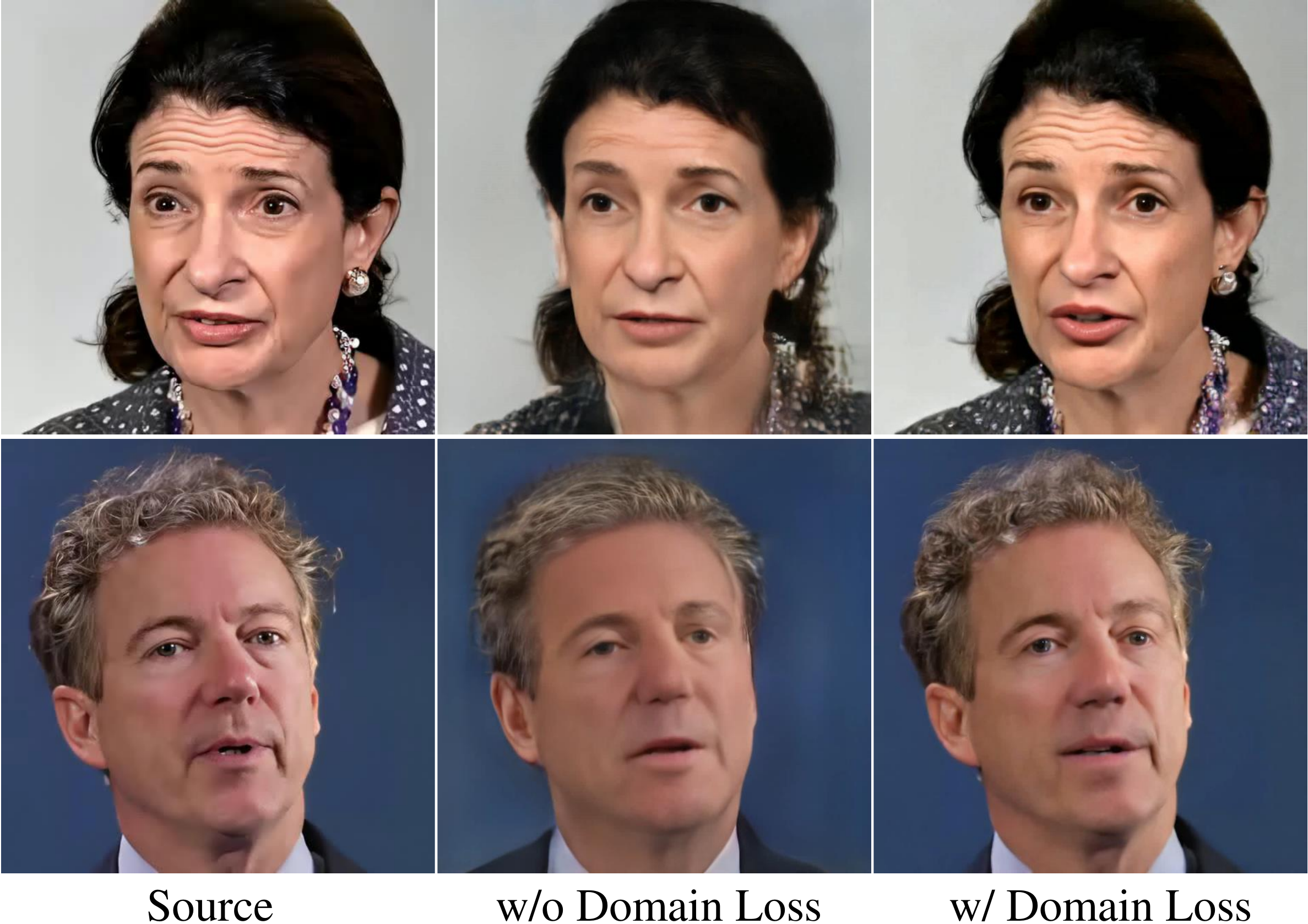}}
\vspace{-0.5em}
\caption{Ablation study of the domain loss.}
\label{fig:loss_ablation}
\end{center}
\vspace{-1em}
\end{figure}

\begin{figure}[t]
\begin{center}
\centerline{\includegraphics[width=1\linewidth]{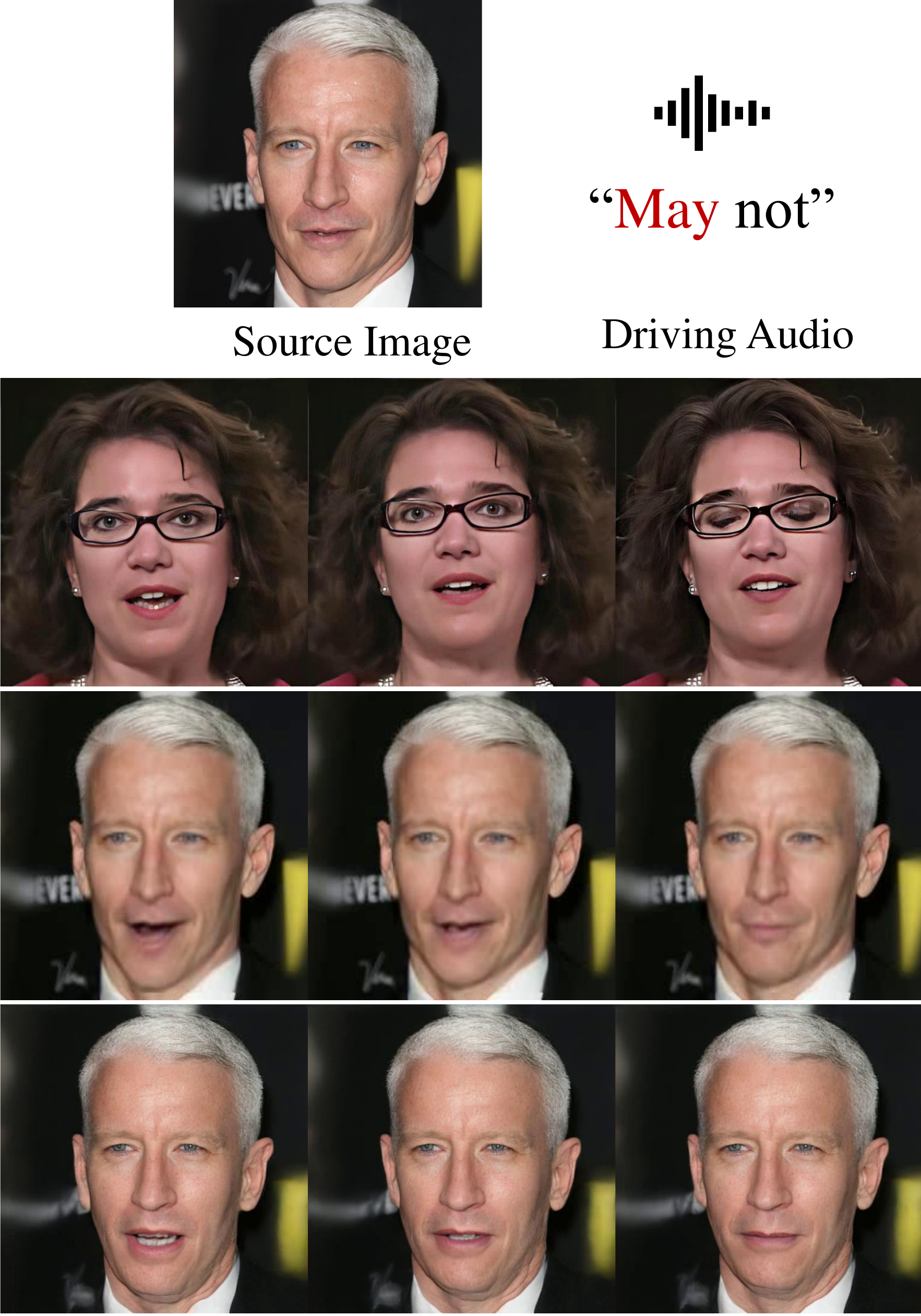}}
\caption{Comparison with wav2lip~\cite{prajwal2020wav2lip}. The audio is used to drive the lip movement generation. 
The second row represents the ground truth frames of the audio segment. 
The third row shows the results generated by wav2lip. 
The bottom shows our results. 
}
\label{fig:audio_wav2lip}
\end{center}
\end{figure}

\subsection{High-Resolution Talking Face Generation} \label{sec:resolution}
As our framework is based on a pre-trained StyleGAN, it can generate high-quality images with a resolution of 1024 $\times$ 1024. 
Currently, there are two existing works that raise the resolution to 512$\times$512, \textit{i.e.,}~\cite{face_vid2vid} and~\cite{zhang2021hdtf}. 
Both of them are trained on newly collected high-resolution datasets, \textit{i.e.,} TalkingHead-1KH and HDTF. 
However, they release neither the source code nor the pre-trained model, which sets an obstacle for us to compare visual quality with them. 
There are many methods that can reach the resolution of 256 $\times$ 256. 
Among them, the recently published PIRenderer~\cite{ren2021pirenderer} and HeadGAN~\cite{doukas2021headgan} stand out due to their satisfying visual quality. 
As PIRenderer provides the source code and the checkpoint, here we compare with it for illustration. 

Fig.~\ref{fig:resolution} shows some visual results of PIRenderer and ours. 
We use a real face to drive a synthetic face or a real face. 
The synthetic images are randomly sampled from StyleGAN. 
It can be observed that both methods can transfer the head pose and facial expression well. 
However, PIRenderer looks blurry and loses some facial details due to the low resolution. 
While our high-resolution results contain more visual details, especially around eyes, mouth, and hair. 
The pupil, wrinkle, splash, and even a single hair can be noticeable. 


Can face restoration techniques promote the low-resolution methods? 
We combine Bi-layer, FOMM, and PIRenderer with a state-of-the-art Blind Face Restoration method, \textit{i.e.,} GFP-GAN~\cite{wang2021gfpgan}, to improve the resolution and image quality. 
GFP-GAN leverages the image prior in StyleGAN to recover facial details. 
It improves the resolution of these methods to 1024$\times$1024. 
The results of these enhanced methods and ours are shown in Fig.~\ref{fig:sr_ablation}. 
We can observe that GFP-GAN greatly improves the visual quality for these low-resolution methods. 
However, it also brings a lot of side effects. 
First, the faces are over smoothed. 
The details on the skin are removed and small wrinkles and the beard are gone. 
The textures of the hair and brow are weakened. 
On the contrary, a single hair or brow is visible in our result. 
Second, the color tone of skin becomes different from the source. 
Our method keeps the color tone better. 
Third, the lightning is also changed compared to the source. 
Moreover, the face restoration cannot remedy the generated artifacts. For example, the artifacts in the mouth generated by PIRenderer are still there after restoration. 
The results demonstrate that our method outperforms the competing methods in terms of image quality even though they are enhanced by the face restoration method.

\subsection{Video-Driven Face Reenactment}\label{sec:motion}
To evaluate the performance of video-driven motion transfer, we conduct two facial reenactment tasks, \textit{i.e.,} same-identity reenactment and cross-identity reenactment.
For the same-identity reenactment, the identity of the source portrait is the same as that of the driving video. 
For cross-identity reenactment, the identity of the source portrait differs from that of the driving video. The latter is more challenging than the former due to the facial shape gap between the source and driving identities.

For the same-identity case, we perform an experiment on 20 selected test videos from the HDTF dataset. 
We treat the first frame as the source portrait and the next 500 frames as the driving video. 
Hence, we have 10,000 synthesized frames of which each has a corresponding ground-truth frame. 

For the cross-identity case, we use the 1000 images of the CelebA-HQ dataset as the source portraits and the 20 test videos of HDTF as driving videos. 
We use the first 100 frames of each driving video to drive 50 source images. 
Hence, we can obtain 100,000 synthesized images. 
As we do not have the ground-truth image in this case, we can only compute FID for the synthetic images, CSIM between the source and synthetic images, and AED and APD between the driving and synthetic images. 

We compare our method with several state-of-the-art methods, including 
X2Face~\cite{wiles2018x2face}, Bi-layer~\cite{zakharov2020bilayer}, FOMM~\cite{siarohin2019fomm}, and PIRenderer~\cite{ren2021pirenderer}. 
All these methods are open-source. 
We use the officially released checkpoints for evaluation. 

\paragraph{Qualitative Evaluation.} 
The visual results of the same-identity and cross-identity are shown in Fig.~\ref{fig:same_id} and Fig.~\ref{fig:cross_id}, respectively. 
Our method can achieve superior image resolution and quality over other methods, which has been illustrated in detail in Sec.~\ref{sec:resolution}. 
Here we focus on other aspects. 
In the same-identity case, all methods perform well in transfer pose except X2Face. 
X2Face suffers from extreme distortion when the pose of the source image differs a lot from the driving image (see the last row of Fig.~\ref{fig:same_id}).
For expression, our method outperforms other methods when there is a large expression difference between the source and driving images, especially when the mouth of the source is closed while that of the driving image is opened by a large margin. 
For instance, in the first and third rows of Fig.~\ref{fig:same_id}, other methods encounter distortions within the mouth. They fail to generate clear teeth while our method works much better. 

In the cross-identity case, more issues occur for other methods while our method can work stably. 
FOMM suffers from head distortion. 
As it is purely based on 2D warping, it can hardly handle the head shape gap between the source and driving images. 
We find PIRenderer is not perfect in handling eye gaze. For instance, in the first row, PIRenderer synthesizes an image with closed eyes, which is not expected compared to the driving image. 
The same problem also appears in the last two rows. 
PIRenderer applies the motion field to the input image and then uses a network for refinement in the image space. 
Differently, leveraging the framework of a pre-trained StyleGAN, we apply the prediction motion to the feature map and design a calibration network to remedy distortion in the feature space. 
The pre-trained parameters convert the calibrated feature maps to a high-quality image. 
We attribute the performance gain over PIRenderer to the powerful image prior in StyleGAN.

\paragraph{Quantitative Evaluation.} Quantitative results of the two reenactment tasks are shown in Table~\ref{tb:criterion}. 
Our FID is the best in the cases, which indicates our synthesized faces are more realistic than those of other methods. 
It is the benefit of raising the image resolution to cover more visual details. 
Our better LPIPS and PSNR mean that we have better reconstruction performance. 
Interestingly, FOMM achieves the best CSIM in the same-identity reenactment, but worse CSIM in the cross-identity reenactment. 
One reason is that FOMM performs well when there is no face shape gap between the source and driving images. 
But when the shape gap is large, it suffers from large head distortion that affects a lot on the identity similarity for a face recognition model. 
As our method uses a GAN inversion method to get the feature maps, it inevitably loses some identity information in the reconstruction. 
This might cause our lower CSIM in the same-identity case. 
On the contrary, our best CSIM in the cross-identity case indicates that our method can work stably in this more challenging setting and suffer from less distortion. 
Our AED and APD are comparable to PIRenderer in the two reenactment tasks.

\subsection{Audio-Driven Talking Face Generation}
In our framework, the audio-based motion generator can work either individually or jointly with the video-based motion generator. 
The visual results of both cases are illustrated in Fig.~\ref{fig:audio}. 
The first row represents the videos that provide the audios. 
The first column represents the source portraits to be animated. 
Synthesized faces from the 2nd to 5th column are generated purely by the driving audio. 
While synthesized faces in the last three columns are generated according to both the driving video and audio. The driving video controls the head pose while the audio controls the lip movement. 

For the audio-driven case, it can be observed that the generated lip movements are consistent with those of the ground-truth video for different source portraits. 
For the audio-and-video-driven case, the results show that the pose is accurately controlled by the video and the lip movements are still consistent with those of the video. 
Both the visual and acoustic control can generalize to different identities.

We also compare with the state-of-the-art audio-driven talking face generation method,  wav2lip~\cite{prajwal2020wav2lip}. 
The visual results are shown in Fig.~\ref{fig:audio_wav2lip}. 
Our results have much better visual quality than wav2lip as wav2lip cannot handle high-resolution input. The mouth of wav2lip is blurred and no teeth are synthesized.


\subsection{Talking Face Video Editing}
\label{sec:editing}

Our framework enables two types of face editing for talking face, \textit{i.e.,} global facial attribute editing and intuitive face editing. 

\paragraph{Global Attribute Editing.}
As our model is built upon a pre-trained StyleGAN, it inherits a powerful property of StyleGAN, \textit{i.e.,} facial attribute editing in the latent style space via existing GAN inversion methods. 
One distinct advantage of our framework is efficiency because we can freely edit facial attributes anytime during the talking video generation with performing GAN inversion only once. 
The attribute editing and the feature editing happen in the same forward process of StyleGAN. 
On the contrary, for other one-shot talking head methods, if they intend to change attributes for each frame, they have to perform the GAN inversion first to obtain the edited image and then perform motion transfer to that image. Hence, they have to perform GAN inversion multiple times for all frames, which is rather time-consuming.  

Our framework is convenient to edit attributes globally. 
We apply the GAN inversion method~\cite{wang2021hfgi} to obtain the latent style codes for the first frame. 
Then, we can freely apply pre-defined style directions to change the style codes with a controllable extent in the video generation process. 
We can efficiently modify the attributes by adding a shift in the style directions to the style codes at any timestamp. 
Fig.~\ref{fig:editing_1} illustrates the visual results of gradually editing several attributes in videos, including makeup, beard, and age. 
It can be observed that attribute editing is stable and does not have side effects on the motion transfer, \textit{i.e.,} attribute editing and motion transfer works with no-interference from each other. 



\paragraph{Intuitive Editing.}

Our video-based motion generator uses 3DMM parameters of the driving image to guide the motion generation for the source image. 
As 3DMM based talking face generation methods~\cite{ren2021pirenderer,doukas2021headgan} always enable the intuitive editing on pose and expression, this also enables us to control the motion generation by directly modifying the 3DMM parameters, resulting in the intuitive editing on the final synthesis. 
We compare with a state-of-the-art 3DMM based method PIRenderer~\cite{ren2021pirenderer}. 
The results are shown in Fig.~\ref{fig:intuitive_control}. 
Both PIRenderer and our method can accurately transfer the pose and expression from 3DMM parameters. 
But thanks to the facial prior preserved in the StyleGAN, we can achieve much better visual quality in terms of resolution, texture, and lightning. 
For instance, in the third row, the lightning on the noise and check of the source portrait is perfectly reserved under different poses and expressions. 
While the lightning of the synthetic images by PIRenderer becomes less noticeable. 


\begin{figure}[ht]
\begin{center}
\centerline{\includegraphics[width=1\linewidth]{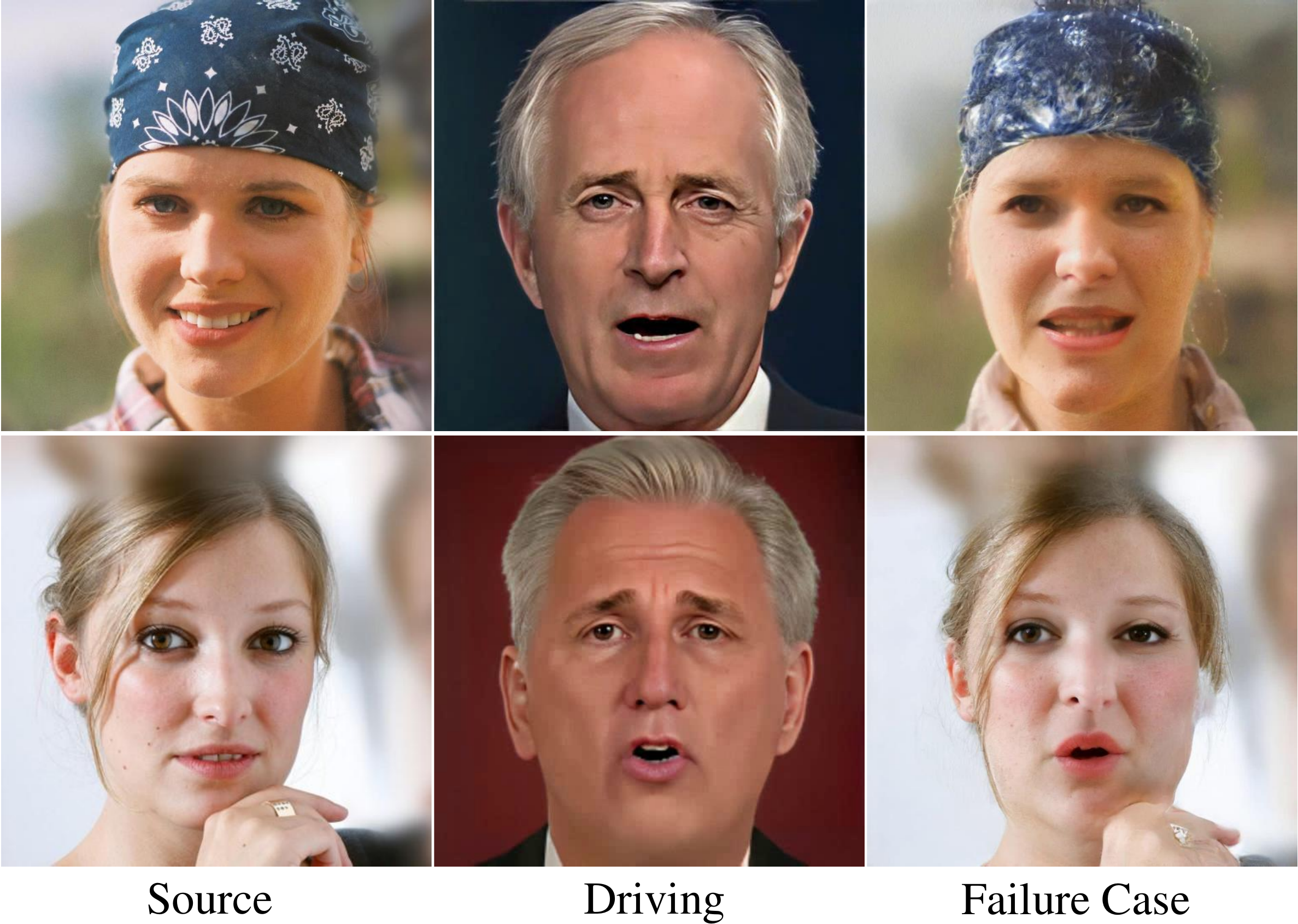}}
\vspace{-0.5em}
\caption{Some failure cases. The headband region (top) and the occluded region by hand (bottom) cannot be well preserved in the synthesized images.}
\label{fig:failure_case}
\end{center}
\end{figure}

\subsection{Ablation Study}
We perform ablation studies to verify the effectiveness of several important designs in our framework that improve the image quality.

\paragraph{Calibration Network.} 
Directly applying the flow fields to the feature map will lead to apparent artifacts around eyes and mouth, \textit{e.g.,} 2D warping is unable to generate teeth for a closed mouth. 
Hence, we design the calibration network to rectify the artifacts caused by warping in the feature space.
We compare the performance with or without the calibration network. 
The results are shown in Fig.~\ref{fig:calibration_ablation}. 
It can be observed that the model cannot correctly generate teeth and closed eyes without using the calibration network. 
The calibration network greatly improves the shape and content around the eyes and mouth.


\paragraph{Domain Loss.} 
The calibration network modifies the feature maps. 
To prevent the edited feature maps from going far away from the original feature maps, we design the domain loss. 
We compare the performance with or without the domain loss. 
The results are shown in Fig.~\ref{fig:loss_ablation}. 
We can observe that dropping the loss makes the synthetic images blurry and lose facial details such as wrinkles and hair texture. 


\section{Limitation and Discussion}
The proposed approach also has some limitations. 
Fig.~\ref{fig:failure_case} shows two types of failure cases. 
First, existing GAN inversion methods cannot be perfect:
reconstruction errors cannot be completely avoided due to the information lost in the feature maps.  
As shown in the first row, the headband texture of the source portrait is severely distorted. 
This type of failure could be alleviated by using better GAN inversion techniques developed in the future.  
Second, our framework currently cannot handle facial occlusions. 
As shown in the second row, there are notable artifacts around the occluded region in the synthetic image. 
This is a common problem for all existing talking face generation methods that need to be addressed in the future. 

As proposed in \cite{stylegan3}, there exist texture-sticking artefacts of images generated by StyleGAN2, which means the hair and face in synthesised videos typically do not move in unison. Alias-Free GAN~\cite{stylegan3} designs a specific architecture to overcome the problem. Our framework can be migrated to the new generator when high-quality GAN inversion methods are studied.

\section{Conclusion}
We propose a novel framework for one-shot talking face generation based on a pre-trained StyleGAN.
The powerful StyleGAN enables a set of powerful functionalities: high-resolution talking video generation (1024$\times$1024 for the first time), disentangled control by driving video and audio, and flexible face editing. 
Our system consists of a video-based motion generation module, an audio-based motion generation module, and a calibration network, which cooperate with the pre-trained StyleGAN to generate high-quality talking faces. 
The framework enables video-driven reenactment, audio-driven reenactment, and video-and-audio jointly driven reenactment. 
Besides, our framework allows two types of face editing, \textit{i.e.,} global attribute editing via GAN inversion and intuitive editing based on 3DMM. 
We conduct comprehensive experiments to illustrate various capabilities of our unified framework, including ablation studies and comparisons with many state-of-the-art methods.


\bibliographystyle{ACM-Reference-Format}
\bibliography{reference.bib}










\end{document}